\documentclass[11pt]{article}
\usepackage[preprint]{acl}
\newif\ifreview
\reviewfalse
\usepackage{times}
\usepackage{latexsym}
\usepackage[T1]{fontenc}
\usepackage[utf8]{inputenc}
\usepackage{microtype}
\usepackage{inconsolata}
\usepackage{graphicx}
\usepackage{xurl}
\usepackage{booktabs}
\usepackage{amsmath}
\usepackage{amssymb}
\usepackage{algorithm}
\usepackage{algpseudocode}
\usepackage{listings}
\usepackage{enumitem}
\usepackage{xcolor}
\usepackage{multirow}
\usepackage{array}
\usepackage{tabularx}
\usepackage{float}
\usepackage{tikz}
\usetikzlibrary{positioning,arrows.meta,calc}
\newcolumntype{Y}{>{\raggedright\arraybackslash}X}

\setlist{itemsep=1pt,topsep=2pt,parsep=0pt}

\newcommand{\taumin}{\tau_{\min}}
\newcommand{\taumax}{\tau_{\max}}
\newcommand{\taumerge}{\tau_{\text{merge}}}
\newcommand{\tauorphan}{\tau_{\text{orphan}}}
\newcommand{\Kmax}{K_{\max}}
\ifreview
  \newcommand{\system}{\textsc{AnonChunker}}
  \newcommand{\sysversion}{\textit{[version withheld for review]}}
\else
  \newcommand{\system}{\textsc{MDKeyChunker}}
  \newcommand{\sysversion}{0.3.0}
\fi
\newcommand{\Estrong}{E\textsubscript{s}}
\newcommand{\CRstrong}{CR\textsubscript{s}}

\input{numbers}

\ifreview
  \newcommand{\repo}{in the anonymized supplementary material}
  \newcommand{\cacherelease}{are released with the public version of the paper (too large for the supplementary material)}
  \newcommand{\commit}[1]{\textit{[hash withheld for review]}}
  
  \newcommand{\Earlierversions}{An earlier preprint (withheld for review)}
\else
  \newcommand{\repo}{at \url{https://github.com/bhavik-mangla/MDKeyChunker}}
  \newcommand{\cacherelease}{are on Hugging Face at \url{https://huggingface.co/datasets/bhavikmangla/mdkeychunker-study}}
  \newcommand{\commit}[1]{\texttt{#1}}
  
  \newcommand{\Earlierversions}{Versions~1 and~2 of this paper on arXiv}
\fi

\title{Upgrade the Retriever First: What an LLM Call per Chunk Buys in RAG}

\author{Bhavik Mangla \\
  Independent Researcher \\
  \texttt{bhavikmangla1234@gmail.com} \\
  Code and data: \url{https://github.com/bhavik-mangla/MDKeyChunker}}

\begin{document}
\maketitle
\begin{abstract}
Many RAG pipelines pay one LLM call per chunk at indexing time to add titles, summaries, keywords and questions (enrichment) or document context (contextual retrieval, CR). We ask what that call buys over free structure and a better retriever. Our advice: upgrade the retriever first, chunk at headers, and enrich mainly for lexical retrieval. Under a plan made public before any index was built, 7B enrichment beats a free section-path prefix under a hybrid retriever by \R{H1-qasper-hyb-d} recall points on Qasper; on FreshStack its gain (\R{H1-fs-hyb-d}) does not survive Holm adjustment. It is equivalent to CR, which sends \R{crratio-local-qasper}--\R{crratio-local-fs} times the input tokens, and Qasper answer F1 (7B reader) is equivalent for both contrasts. Header-led chunks beat 512-character windows; on FreshStack they lose to overlapping 256-token windows. Under qwen3-embedding and a reranker, the two strongest of eight retrievers (named as secondary before any outcome was computed), the CIs of 7B enrichment against the free prefix fall within the equivalence margins; under BM25 it gains \R{H1-qasper-bm25}. Exploratory analyses attribute the hybrid gain to document expansion and to repairing an untuned rank fusion; the free prefix under either strong retriever beats hybrid enrichment by \R{ax-fr-fs-STrr-Ehyb-d} to \R{ax-fr-qasper-STrr-Ehyb-d} points, though the best 120B-indexed configuration leads the best free one by \R{ax-fr-qasper-best-bestfree-d} on Qasper. Add-ons built on the call do no better than LLM-free controls, and late chunking lowers Qasper recall. We release the plan, harness and per-question results.
\end{abstract}

\section{Introduction}
\label{sec:intro}

\begin{figure*}[t]
\centering
\begin{tikzpicture}[font=\scriptsize]
\draw (0.00,0) rectangle (4.60,3.60);
\node[above] at (2.30,3.60) {Qasper (rec@512t)};
\draw[gray!25] (0.00,0.000) -- (4.60,0.000);\node[left] at (0.00,0.000) {25};
\draw[gray!25] (0.00,0.514) -- (4.60,0.514);\node[left] at (0.00,0.514) {30};
\draw[gray!25] (0.00,1.029) -- (4.60,1.029);\node[left] at (0.00,1.029) {35};
\draw[gray!25] (0.00,1.543) -- (4.60,1.543);\node[left] at (0.00,1.543) {40};
\draw[gray!25] (0.00,2.057) -- (4.60,2.057);\node[left] at (0.00,2.057) {45};
\draw[gray!25] (0.00,2.571) -- (4.60,2.571);\node[left] at (0.00,2.571) {50};
\draw[gray!25] (0.00,3.086) -- (4.60,3.086);\node[left] at (0.00,3.086) {55};
\draw[gray!25] (0.00,3.600) -- (4.60,3.600);\node[left] at (0.00,3.600) {60};
\node[below, align=center] at (0.460,0) {free\\[-1pt]{\color{gray}0}};
\node[below, align=center] at (1.380,0) {E\\[-1pt]{\color{gray}1.0k}};
\node[below, align=center] at (2.300,0) {CR\\[-1pt]{\color{gray}5.9k}};
\node[below, align=center] at (3.220,0) {E$_{s}$\\[-1pt]{\color{gray}1.1k}};
\node[below, align=center] at (4.140,0) {CR$_{s}$\\[-1pt]{\color{gray}6.3k}};
\draw[densely dotted] (0.460,0.503) -- (1.380,1.398) -- (2.300,0.885) -- (3.220,1.534) -- (4.140,0.899);
\fill (0.460,0.503) circle (1.1pt);
\fill (1.380,1.398) circle (1.1pt);
\fill (2.300,0.885) circle (1.1pt);
\fill (3.220,1.534) circle (1.1pt);
\fill (4.140,0.899) circle (1.1pt);
\draw[solid] (0.460,1.817) -- (1.380,2.381) -- (2.300,2.167) -- (3.220,2.502) -- (4.140,2.339);
\fill (0.460,1.817) circle (1.1pt);
\fill (1.380,2.381) circle (1.1pt);
\fill (2.300,2.167) circle (1.1pt);
\fill (3.220,2.502) circle (1.1pt);
\fill (4.140,2.339) circle (1.1pt);
\draw[dashed] (0.460,2.942) -- (1.380,3.098) -- (2.300,3.156) -- (3.220,3.041) -- (4.140,3.200);
\fill (0.460,2.942) circle (1.1pt);
\fill (1.380,3.098) circle (1.1pt);
\fill (2.300,3.156) circle (1.1pt);
\fill (3.220,3.041) circle (1.1pt);
\fill (4.140,3.200) circle (1.1pt);
\draw[very thick] (0.460,3.030) -- (1.380,3.070) -- (2.300,3.038) -- (3.220,3.287) -- (4.140,3.287);
\fill (0.460,3.030) circle (1.1pt);
\fill (1.380,3.070) circle (1.1pt);
\fill (2.300,3.038) circle (1.1pt);
\fill (3.220,3.287) circle (1.1pt);
\fill (4.140,3.287) circle (1.1pt);
\draw (5.80,0) rectangle (10.40,3.60);
\node[above] at (8.10,3.60) {FreshStack (prec@1024t)};
\draw[gray!25] (5.80,0.000) -- (10.40,0.000);\node[left] at (5.80,0.000) {20};
\draw[gray!25] (5.80,0.900) -- (10.40,0.900);\node[left] at (5.80,0.900) {25};
\draw[gray!25] (5.80,1.800) -- (10.40,1.800);\node[left] at (5.80,1.800) {30};
\draw[gray!25] (5.80,2.700) -- (10.40,2.700);\node[left] at (5.80,2.700) {35};
\draw[gray!25] (5.80,3.600) -- (10.40,3.600);\node[left] at (5.80,3.600) {40};
\node[below, align=center] at (6.375,0) {free\\[-1pt]{\color{gray}0}};
\node[below, align=center] at (7.525,0) {E\\[-1pt]{\color{gray}1.0k}};
\node[below, align=center] at (8.675,0) {CR\\[-1pt]{\color{gray}8.8k}};
\node[below, align=center] at (9.825,0) {E$_{s}$\\[-1pt]{\color{gray}1.1k}};
\draw[densely dotted] (6.375,0.489) -- (7.525,0.919) -- (8.675,0.542) -- (9.825,0.875);
\fill (6.375,0.489) circle (1.1pt);
\fill (7.525,0.919) circle (1.1pt);
\fill (8.675,0.542) circle (1.1pt);
\fill (9.825,0.875) circle (1.1pt);
\draw[solid] (6.375,1.442) -- (7.525,1.935) -- (8.675,1.620) -- (9.825,1.761);
\fill (6.375,1.442) circle (1.1pt);
\fill (7.525,1.935) circle (1.1pt);
\fill (8.675,1.620) circle (1.1pt);
\fill (9.825,1.761) circle (1.1pt);
\draw[dashed] (6.375,2.620) -- (7.525,2.667) -- (8.675,2.773) -- (9.825,2.833);
\fill (6.375,2.620) circle (1.1pt);
\fill (7.525,2.667) circle (1.1pt);
\fill (8.675,2.773) circle (1.1pt);
\fill (9.825,2.833) circle (1.1pt);
\draw[very thick] (6.375,2.546) -- (7.525,2.688) -- (8.675,2.525) -- (9.825,2.861);
\fill (6.375,2.546) circle (1.1pt);
\fill (7.525,2.688) circle (1.1pt);
\fill (8.675,2.525) circle (1.1pt);
\fill (9.825,2.861) circle (1.1pt);
\draw[densely dotted] (0.20,-1.0) -- (0.80,-1.0);\node[anchor=west] at (0.85,-1.0) {BM25};
\draw[solid] (2.80,-1.0) -- (3.40,-1.0);\node[anchor=west] at (3.45,-1.0) {Hybrid};
\draw[dashed] (5.40,-1.0) -- (6.00,-1.0);\node[anchor=west] at (6.05,-1.0) {qwen3};
\draw[very thick] (8.00,-1.0) -- (8.60,-1.0);\node[anchor=west] at (8.65,-1.0) {Hybrid+rerank};
\end{tikzpicture}
\caption{\textbf{Paid indexing lifts Qasper retrieval under BM25 and the default hybrid, and, with the 7B indexing LLM, barely moves it under qwen3-embedding or a reranker.} Mean primary metric (Qasper rec@512t; FreshStack prec@1024t) per indexing arm (x-axis) and retriever (lines). Arms: free section-path prefix (free; S+T), enrichment (E), contextual retrieval (CR) and their 120B-indexed versions (\Estrong{}, \CRstrong{}; \CRstrong{} Qasper only); grey numbers are indexing-LLM input tokens per chunk. The free prefix under qwen3 or the reranker scores above every paid arm under the hybrid. Paired differences with CIs: Table~\ref{tab:confirm}, Figure~\ref{fig:retrievers}.}
\label{fig:frontier}
\end{figure*}

Retrieval-augmented generation (RAG) indexes documents as chunks~\citep{lewis2020rag}. Two choices shape what the retriever can find: where the chunk boundaries fall, and what text is indexed for each chunk~\citep{chen2024densex}. Some of this signal is free. Markdown, HTML and many converted documents mark their sections with headers, so a parser can cut at section boundaries and prefix each chunk with its section path~\citep{yang2026titlechain}. Other signal costs one or more model calls per chunk at indexing time: questions the chunk answers~\citep{nogueira2019doc2query}, a sentence that places the chunk in its document~\citep{anthropic2024contextual}, or titles, summaries and keywords~\citep{mishra2025metadata}.

Teams that build RAG must decide whether that call is worth its cost, and whether the same budget would do more if spent on the retriever. We ask three questions. \emph{What does one LLM call per chunk add to free document structure?} We call this step \emph{enrichment} (E) and compare it with a free section-path prefix and with contextual retrieval (CR)~\citep{anthropic2024contextual}, which sends the whole document with every chunk. \emph{Does a stronger retriever change the answer, and why?} \emph{Do the add-ons that pipelines build on the call earn their cost?} Such add-ons reuse labels~\citep{pham2024topicgpt}, group chunks with an LLM~\citep{duarte2024lumberchunker,zhao2024metachunking} or link related passages~\citep{edge2024graphrag,gutierrez2024hipporag}. Our answer is conditional: enrichment pays on Qasper under a default hybrid retriever, but under the two strongest retrievers we tested, enrichment with the 7B indexing LLM adds nothing detectable, and in an exploratory contrast the free prefix under qwen3-embedding beats hybrid enrichment by \R{ax-fr-qasper-STq3-Ehyb-d} points on Qasper. For such corpora, the first money is better spent on the retriever.

Our contributions are:
\begin{enumerate}[leftmargin=*,label=(\roman*)]
  \item \textbf{A confirmatory test with matched free and paid controls.} A plan made public before the first build fixed the hypotheses, metrics, retriever, equivalence margins and multiplicity families (\S\ref{sec:plan}). On 951 Qasper and 235 FreshStack questions, with a 7B and a 120B indexing LLM, enrichment beats the free prefix under the primary hybrid retriever on Qasper by \R{H1-qasper-hyb-d} points; the FreshStack gain (\R{H1-fs-hyb-d}) does not survive Holm adjustment~\citep{holm1979simple}. Enrichment is equivalent to CR, which sends \R{crratio-local-qasper}--\R{crratio-local-fs} times the input tokens per call, and Qasper answer F1 is equivalent for both contrasts. Header-led chunks beat 512-character windows on both datasets, but 256-token overlapping windows beat them on FreshStack.
  \item \textbf{Stronger retrievers leave enrichment little to add.} With the 7B indexing LLM, enrichment is equivalent to the free prefix under qwen3-embedding and under a reranker on both datasets, and to CR under the reranker (Table~\ref{tab:confirm}). Both retrievers were named as secondary before any outcome was computed. Exploratory analyses suggest why the hybrid gains: it behaves as doc2query-style document expansion~\citep{nogueira2019doc2query}, and much of it repairs an untuned rank fusion, since hybrid enrichment on Qasper scores \R{m-qasper-hybrid-E} and the free prefix under nomic-embed-text alone \R{m-qasper-nomic-S+T}.
  \item \textbf{Add-ons built on the call do not earn their cost.} Pre-specified add-on tests compare a rolling key dictionary, key-based merging and related-key expansion with LLM-free controls. None beats its LLM-free control; one call for all fields matches one call per field.
  \item \textbf{A spending guide and released materials} (\S\ref{sec:discussion}). We release the harness, the plan and its amendments, and per-question results.
\end{enumerate}

\section{Related Work}
\label{sec:related}

\paragraph{Is costly chunking worth it?}
\citet{caspari2026complex} is the closest study. Across eight strategies, including contextual, summary-enriched and late chunking, they find that costly methods rarely give consistent gains, and that cheap title enrichment is competitive. They use dense retrievers only; we add header-structured documents, lexical and hybrid retrieval, equivalence tests, CR at the same number of calls, and whether retriever strength changes the answer. Other studies find that the optimum depends on the task and that costly segmenters rarely pay off~\citep{zhou2026beyond,smigielski2026chunking,qu2025semantic,smith2024chroma}. \citet{shaukat2026systematic} report that content-aware chunking beats fixed windows, a comparison open to the alignment confound we control for (\S\ref{sec:arms}). HiChunk~\citep{lu2025hichunk} shows that existing benchmarks are evidence-sparse; we therefore match evidence at the paragraph or span level under a fixed token budget. Concurrently, \citet{kuehlkamp2026structure} run pre-registered, placebo-controlled contrasts with dense embedders on Wikipedia and on Qasper pooled across papers: structure-aligned chunks with real heading paths beat placebo paths shuffled across documents and LLM-contextualized fixed windows. We add lexical and hybrid retrieval, equivalence tests, metadata enrichment and cost per call. Within single papers the prefix itself has no detectable effect (\R{STS-qasper}, exploratory), plausibly because heading paths help most when they separate documents.

\paragraph{Free structure.}
\citet{yang2026titlechain} reports that a three-stage pipeline (header splits, semantic merging and a title-chain prefix) improves MRR@5 by 23.8\% on a production knowledge base, without isolating the prefix. Its gold labels are LLM-judged chunk-level scores; we test the prefix alone, with gold evidence defined on the source text, independent of any chunking (\S\ref{sec:metrics}).

\paragraph{Paid text at index time.}
Doc2query~\citep{nogueira2019doc2query} appends predicted questions to each chunk; later work filters them~\citep{gospodinov2023doc2queryminus} or fuses two indexes~\citep{kuo2025doc2queryplus}. Expansions help weak retrievers and hurt strong ones~\citep{weller2024expansions}; we observe the first half and, under strong retrievers, equivalence rather than a loss (\S\ref{sec:why}). \citet{mishra2025metadata} and \citet{dadopoulos2025metadata} report gains from LLM metadata without a CR or free-prefix control. \citet{reuter2025sac} prepend one document summary to every chunk to reduce retrieval from the wrong document in legal corpora. \citet{zerhoudi2026metadata} find that adding metadata to the answering model's context mostly lowers accuracy. CR~\citep{anthropic2024contextual} gives the model the whole document for each chunk; late chunking~\citep{gunther2024late} pools token embeddings of the whole document without any LLM call. ConTEB~\citep{conti2025conteb} shows that embedders fail when context is needed. \citet{merola2025reconstructing} find CR better than late chunking but costlier. Appendix~\ref{app:comparison} compares how these methods spend calls.

\paragraph{Label reuse, grouping and context expansion.}
TopicGPT~\citep{pham2024topicgpt} shows an LLM the existing topic list so that it reuses labels, and merges near-duplicates in a later pass. It works across a corpus, keeps no cap and does not feed chunking. Our rolling dictionary applies label reuse within one document, and our consolidation control (H7, \S\ref{sec:mech}) follows the TopicGPT pattern. LumberChunker~\citep{duarte2024lumberchunker} and Meta-Chunking~\citep{zhao2024metachunking} place boundaries with an LLM or with perplexity, but form only contiguous chunks. RAPTOR~\citep{sarthi2024raptor} groups non-adjacent chunks by embedding clusters; our key-based merging (\S\ref{sec:arms}) does so by an LLM-assigned label, so we compare it with count-matched random and similarity merges. At retrieval time, HiChunk~\citep{lu2025hichunk} merges sibling chunks into their parent within a token budget, and neighbour-window and parent-document expansion are common baselines. Graph-based RAG follows links between entities or passages~\citep{edge2024graphrag,gutierrez2024hipporag}. We compare related-key links with equal-token neighbour and section expansion.

\section{Indexing Arms}
\label{sec:method}
\label{sec:arms}
\emph{Header-led chunking} reads a document into blocks (headers, code, tables, lists, paragraphs) and tracks each block's section path, the chain of enclosing header titles. A header starts a new chunk; within a section, blocks are packed up to $\taumax{=}1{,}500$ characters, and no block is split. Chunks under $\taumin{=}100$ characters join a neighbour. \emph{Enrichment} (E) makes one LLM call per chunk. It returns a title, a summary, 5--8 keywords and 2--3 questions the chunk answers, which are indexed before the chunk text. The prompt holds the chunk, its section path, its position and the previous chunk's summary. CR instead gives the model the whole document and asks for a short context that situates the chunk.

We use the open-source implementation \system{}\footnote{\Earlierversions{} reported an evaluation of this system that we found invalid and do not use: its full-pipeline configuration never indexed the generated metadata, and its relevance labels were hand-written substring matches (Appendix~\ref{app:withdrawn}).} (Appendix~\ref{app:method}, Figure~\ref{fig:pipeline}). Its call also returns a short subtopic \emph{key} and up to three \emph{related keys}. They are not indexed, but they drive three optional add-ons, each tested against an LLM-free control (\S\ref{sec:mech}): a \emph{rolling key dictionary} lists the document's earlier keys in the prompt so that the model can reuse one (arm E uses it); \emph{key-based merging} concatenates a document's chunks that share a key; and \emph{related-key expansion} adds linked chunks at retrieval time. Total enrichment input grows linearly with document length; uncached CR input grows quadratically (Appendix~\ref{app:cost}).

Table~\ref{tab:arms} lists the arms. S indexes the chunk text alone, and S+T prefixes it with its section path (the free prefix). S, S+T, E, CR and their 120B versions share the header-led boundaries and differ only in indexed text, so comparing them isolates the text added at indexing time. F512c and F256t are the usual fixed-size baselines. Two controls test whether a structure gain only reflects alignment with the gold unit: P packs whole paragraphs and ignores headers, and FL uses token windows of the median S length. The 7B indexing LLM is \texttt{qwen2.5:7b} (4-bit Q4\_K\_M, Ollama); the 120B indexing LLM is \texttt{gpt-oss-120b} (OpenRouter, provider pinned, low reasoning effort). All LLM calls use temperature~0 and seed~0, and prompts and models were hash-frozen before any build (Appendix~\ref{app:repro}).

\begin{table}[t]
\centering
\scriptsize
\setlength{\tabcolsep}{3pt}
\begin{tabularx}{\linewidth}{@{}l Y c@{}}
\toprule
\textbf{Arm} & \textbf{Boundaries; indexed text} & \textbf{Calls} \\
\midrule
F512c & 512-character windows; text & 0 \\
F256t & 256-token windows, 32 overlap; text & 0 \\
P$^{\circ}$ & paragraphs packed to $\taumax$, headers ignored; text & 0 \\
FL$^{\circ}$ & token windows of median S length; text & 0 \\
S & header-led; text & 0 \\
S+T & header-led; section path + text & 0 \\
E & header-led; fields (7B) + text & 1 \\
CR & header-led; CR context (7B) + text & 1 \\
\Estrong{} & header-led; fields (120B) + text & 1 \\
\CRstrong{} & header-led; CR context (120B) + text (Qasper) & 1 \\
\midrule
\multicolumn{3}{@{}l}{\emph{Add-on arms (Amendments 2--4, \S\ref{sec:mech})}} \\
M & E + key-based merging & 1 \\
RM / SM & E + random / similarity merges, count-matched & 1 \\
E$-$K & E with the key dictionary hidden & 1 \\
M$-$TP & E$-$K + key consolidation per document + key-based merging & 1+ \\
E+KX / AX / SX & E + key-linked / adjacent / same-section expansion & 1 \\
PF & one call per indexed field (100 papers) & 4 \\
\bottomrule
\end{tabularx}
\caption{Arms. Evidence is always matched against chunk text only. Calls are LLM calls per header-led chunk at indexing time. $^{\circ}$Alignment controls. $_s$: 120B indexing LLM.}
\label{tab:arms}
\end{table}

\section{Experimental Design}
\label{sec:design}

\subsection{Plan and evidence tiers}
\label{sec:plan}
The confirmatory plan~\citep{vanmiltenburg2021prereg} fixed the hypotheses, metrics, primary retriever, equivalence margins and multiplicity families. It was pushed to a public repository before the first index was built; it was not externally registered. Amendment~1, pushed while builds ran and before any outcome metric was computed, added secondary retrievers and outcomes. Amendments 2--4 added the add-on tests (\S\ref{sec:mech}) after the H1--H3 results had been viewed; each was pushed before the arms it governs were built or scored. A pilot on 30 Qasper papers and 73 FreshStack questions (exploratory, because its plan was pushed after its results) found every hybrid enrichment comparison inconclusive and set the sample sizes and margins. The confirmatory samples exclude the pilot's papers and questions, and no outcome metric had been computed on them when the plan was written. We report three evidence tiers: \emph{confirmatory} (the plan's Holm families, \S\ref{sec:stats}), \emph{pre-specified secondary} (retrievers, arms and tests that the plan or its amendments name as secondary) and \emph{exploratory} (everything else, including every contrast across retrievers). Neither the plan nor Amendment~1 names equivalence tests under the secondary retrievers; there we apply the plan's H1/H2 margins to unadjusted 95\% CIs. Appendix~\ref{app:plan} gives commits, timestamps, what the pilot informed, and deviations.

\subsection{Datasets}
\label{sec:data}

\paragraph{Qasper.}
Qasper~\citep{dasigi2021qasper} questions were written by annotators who saw only a paper's title and abstract; other annotators answered them and marked evidence paragraphs. We sampled 300 dev and test papers (seed 2027, pilot papers excluded) and kept questions with text evidence: 951 questions with 2,245 evidence paragraphs. Papers are converted to Markdown with nested section headers; header-led chunking yields 7,222 chunks over the 300 papers (24 per paper on average). Retrieval runs within each paper, so the candidate pool is one paper's chunks.

\paragraph{FreshStack.}
FreshStack~\citep{thakur2025freshstack} pairs Stack Overflow questions with documentation corpora. Its gold labels are corpus chunks that GPT-4o judged to support answer nuggets. For Laravel, Angular and YOLO, we identified the repository commit whose documentation matches FreshStack's corpus byte-exactly, and kept gold spans in Markdown files under the docs root. Without the 73 pilot questions, 235 questions remain (89 Laravel, 99 Angular, 47 YOLO). Laravel is indexed in full (98 files, 3,860 header-led chunks); Angular and YOLO are indexed with all gold files plus seeded random distractors, half as many as the gold files (a cap fixed in advance). Retrieval is pooled within each domain.

\subsection{Retrievers}
The primary retriever is \textbf{hybrid}: reciprocal rank fusion~\citep{cormack2009rrf} ($k{=}60$, untuned) of BM25~\citep{robertson2009bm25} and nomic-embed-text~\citep{nussbaum2024nomic}. The plan names BM25 alone, mxbai-embed-large and qwen3-embedding-0.6B (qwen3;~\citealp{zhang2025qwen3embedding}) as secondary retrievers. Amendment~1 adds a BM25+qwen3 hybrid, Qwen3-Reranker-0.6B over the hybrid's top 30 (run over no other retriever; Appendix~\ref{app:secondary}), and late chunking with jina-embeddings-v3~\citep{sturua2024jinav3,gunther2024late}. Nomic alone is exploratory. qwen3's context and truncation settings were fixed after H1--H3 were viewed, before any qwen3 statistic (Appendix~\ref{app:plan}). A new embedder requires re-embedding the corpus; the reranker reuses the existing index.

\begin{table*}[t]
\centering
\footnotesize
\setlength{\tabcolsep}{4pt}
\begin{tabular}{@{}lllcccc@{}}
\toprule
 & & & \multicolumn{2}{c}{\textbf{Qasper} (rec@512t; $\pm$5)} & \multicolumn{2}{c}{\textbf{FreshStack} (prec@1024t; $\pm$4)} \\
\cmidrule(lr){4-5}\cmidrule(lr){6-7}
\textbf{H} & \textbf{A $-$ B} & \textbf{Retriever} & \textbf{Diff.} [CI] & \textbf{Verdict} & \textbf{Diff.} [CI] & \textbf{Verdict} \\
\midrule
\multicolumn{7}{@{}l}{\emph{Confirmatory: primary family (TOST, 90\% CI)}} \\
H1 & E $-$ S+T & hybrid & \R{H1-qasper-hyb} & \R{H1-qasper-hyb-v} & \R{H1-fs-hyb} & \R{H1-fs-hyb-v}$^{\dagger}$ \\
H2 & E $-$ CR & hybrid & \R{H2-qasper-hyb} & \R{H2-qasper-hyb-v} & \R{H2-fs-hyb} & \R{H2-fs-hyb-v} \\
\multicolumn{7}{@{}l}{\emph{Confirmatory: H3 family (two-sided, 95\% CI)}} \\
H3a & S $-$ F512c & hybrid & \R{H3a-qasper-hyb} & \R{H3a-qasper-hyb-v} & \R{H3a-fs-hyb} & \R{H3a-fs-hyb-v} \\
H3b & S $-$ F256t & hybrid & \R{H3b-qasper-hyb} & \R{H3b-qasper-hyb-v} & \R{H3b-fs-hyb} & \R{H3b-fs-hyb-v} \\
\multicolumn{7}{@{}l}{\emph{Confirmatory: key-secondary family (TOST, 90\% CI)}} \\
H1$_s$ & \Estrong{} $-$ S+T & hybrid & \R{H1s-qasper-hyb} & \R{H1s-qasper-hyb-v} & \R{H1s-fs-hyb} & \R{H1s-fs-hyb-v} \\
H2$_s$ & \Estrong{} $-$ \CRstrong{} & hybrid & \R{H2s-qasper-hyb} & \R{H2s-qasper-hyb-v} & -- & -- \\
F1 & E $-$ S+T (answer F1) & hybrid & \R{F1-H1} & \R{F1-H1-v} & -- & -- \\
F1 & E $-$ CR (answer F1) & hybrid & \R{F1-H2} & \R{F1-H2-v} & -- & -- \\
\multicolumn{7}{@{}l}{\emph{Secondary retrievers, 7B indexing LLM (unadjusted 95\% CI; verdicts use the plan's margins)}} \\
H1 & E $-$ S+T & qwen3-emb. & \R{q3H1-qasper} & \R{q3H1-qasper-v} & \R{q3H1-fs} & \R{q3H1-fs-v} \\
H1 & E $-$ S+T & hybrid+rerank & \R{rrH1-qasper} & \R{rrH1-qasper-v} & \R{rrH1-fs} & \R{rrH1-fs-v} \\
H2 & E $-$ CR & hybrid+rerank & \R{rrH2-qasper} & \R{rrH2-qasper-v} & \R{rrH2-fs} & \R{rrH2-fs-v} \\
\bottomrule
\end{tabular}
\caption{\textbf{Enrichment beats the free prefix only on Qasper under the hybrid, matches CR, and is equivalent to the free prefix under qwen3-embedding and the reranker.} Main tests on 951 Qasper and 235 FreshStack questions, with equivalence margins in the header. Paired differences in points with BCa cluster-bootstrap CIs. Confirmatory verdicts follow \S\ref{sec:stats} within each Holm family; in TOST rows \emph{equiv.}\ takes precedence, and \emph{diff.}\ is judged on the 95\% CI (not shown); every secondary-retriever CI lies inside the margin. $_s$: 120B indexing LLM (\texttt{gpt-oss-120b}); \CRstrong{} was built for Qasper only. Answer F1 uses a fixed 7B reader and a $\pm$5 F1 margin. rerank: Qwen3-Reranker-0.6B over the hybrid top-30. $^{\dagger}$\emph{diff.}\ by the plan's rule (95\% CI \R{H1-fs-hyb-ci95} excludes 0); not significant after Holm adjustment (sign-flip $p$ \R{H1-fs-hyb-pperm}). On Qasper F512c and F256t both score \R{m-qasper-F512c}.}
\label{tab:confirm}
\end{table*}

\subsection{Metrics}
\label{sec:metrics}
Chunk sets differ in chunk count and length, so all primary metrics are \emph{budgeted}. Chunks are taken in rank order until $B$ \texttt{cl100k\_base} tokens of \emph{source text} are filled. Generated text, prefixes and CR context are indexed but neither counted nor matched, so indexed text can change a chunk's rank but not whether it contains evidence. Evidence is matched on word 5-grams taken within each retrieved chunk, never across chunk boundaries.
\textbf{Qasper:} evidence recall within 512 tokens (rec@512t). An evidence paragraph counts as retrieved if at least half of its 5-grams are in the selected text.
\textbf{FreshStack:} gold-token precision within 1,024 tokens (prec@1024t), the share of selected 5-grams that lie in gold spans. Gold spans are long documentation sections, so this metric rewards landing in a labelled section more than pinpointing the exact text.

\subsection{Statistics}
\label{sec:stats}
Differences are paired, A $-$ B, in points of the primary metric (one point is 0.01). Intervals come from a bias-corrected and accelerated (BCa) cluster bootstrap (10,000 resamples; clusters are papers on Qasper and questions, stratified by domain, on FreshStack).
\textbf{Families.} H1 (E vs.\ S+T) and H2 (E vs.\ CR), under hybrid retrieval with the 7B indexing LLM on each dataset, form the primary family of four two one-sided tests (TOST;~\citealp{lakens2017equivalence}). The equivalence margins are $\pm$5 points (Qasper) and $\pm$4 points (FreshStack), chosen for power: from the pilot's paired SDs, they give power $\ge$0.85 when the true difference is zero. H3 (S vs.\ F512c and S vs.\ F256t, both datasets) is a family of four two-sided tests. The key-secondary family contains H1 and H2 with the 120B indexing LLM and Qasper answer F1 for E vs.\ S+T and E vs.\ CR (TOST, margin $\pm$5 F1). Each family is Holm-adjusted. Appendix~\ref{app:components} defines the add-on families (Amendments 2--4). Other results carry unadjusted 95\% CIs and are secondary or exploratory.

\textbf{Decision rules.} A TOST row is \emph{equiv.}\ if its Holm-adjusted TOST $p<0.05$, otherwise \emph{diff.}\ if its 95\% CI excludes 0, otherwise \emph{incon.} A two-sided row is \emph{diff.}\ if its 95\% CI excludes 0 and \emph{n.d.}\ (no difference detected) otherwise. In the H3 family and the add-on families (Appendix~\ref{app:components}), \emph{diff.}\ also requires a Holm-adjusted cluster sign-flip permutation $p<0.05$; in the primary and key-secondary families it does not (the plan's rule), and we report that $p$ alongside. Tables show 90\% CIs for TOST rows and 95\% CIs otherwise. When an \emph{equiv.}\ row's 95\% CI excludes 0, we report a small difference within the margin.

\subsection{Add-on tests}
\label{sec:mech}
Amendments 2--4 added add-on tests (pre-specified secondary) that reuse the cached E outputs, the hybrid retriever, the primary metrics and margins, and the rules above. Each tests an add-on against a matched control: key-based merging (M) against count-matched random (RM) and similarity (SM) merges of adjacent chunks, and against no merging (H4a--c); the key dictionary against hiding it (E$-$K, H4d); in-call keys against a TopicGPT-style consolidation call after the fact, the one control that makes an LLM call~\citep[M$-$TP, H7;][]{pham2024topicgpt}; related-key expansion (E+KX) against inserting approximately the same number of adjacent chunks (E+AX) at equal tokens (H5a--b); and one call against one call per field on 100 Qasper papers (PF, H6). Appendix~\ref{app:components} gives the definitions.

\subsection{Answer-level outcomes}
\textbf{Qasper:} answer token F1 (official evaluator, maximum over gold answers). A fixed reader, \texttt{qwen2.5:7b} at temperature 0, receives exactly 512 tokens of retrieved source text.
\textbf{FreshStack} (secondary, added by Amendment~1): strict nugget recall~\citep{pradeep2024autonuggetizer}. The same reader receives 1,024 tokens. The judge, \texttt{llama-3.3-70b-instruct}, is from a different model family than the indexing LLMs and the reader. We accept the judge only if it gives the accepted Stack Overflow answer $\ge$85\% nugget recall and another question's answer $\le$5\%. LLM judges and human assessors often disagree~\citep{thakur2025support,clarke2024llmrelevance}; without a human spot-check ($\kappa\ge0.6$), these results are exploratory.

\section{Results}
\label{sec:results}

\subsection{Confirmatory results}
\label{sec:confirm}

\paragraph{Enrichment against the free prefix (H1).}
On Qasper (Table~\ref{tab:confirm}) enrichment gains \R{H1-qasper-hyb-d} points over the section-path prefix; its 90\% CI \R{H1-qasper-hyb-ci} crosses the $+$5 margin, so equivalence is not shown, and its 95\% CI \R{H1-qasper-hyb-ci95} excludes 0, so the difference is confirmed (Holm-adjusted sign-flip $p<0.001$). On FreshStack the gain is \R{H1-fs-hyb-d} points; its 95\% CI \R{H1-fs-hyb-ci95} excludes 0, the plan's rule for \emph{diff.}, but the Holm-adjusted sign-flip $p$ is \R{H1-fs-hyb-pperm}, so the gain is not significant after Holm adjustment and we do not treat it as confirmed. By domain (secondary), only Laravel's CI excludes 0 (\R{H1-laravel}). The 120B indexing LLM repeats the Qasper gain (\R{H1s-qasper-hyb-d}; FreshStack inconclusive).

\paragraph{Enrichment against contextual retrieval (H2).}
Enrichment is equivalent to CR on both datasets (the point estimates favour E) and with the 120B indexing LLM on Qasper. CR sent \R{tok-local-cr-qasper-percall} (Qasper) and \R{tok-local-cr-fs-percall} (FreshStack) input tokens per call, against about \R{tok-local-enr-qasper-percall} for enrichment. With the 120B LLM, a CR call cost about twice as much as an enrichment call, not \R{crratio-strong-qasper} times (Appendix~\ref{app:cost}).

\paragraph{Structure against fixed windows (H3).}
Header-led chunks beat 512-character windows on both datasets (Qasper \R{H3a-qasper-hyb-d}, FreshStack \R{H3a-fs-hyb-d}). Against 256-token windows with overlap, S wins on Qasper (\R{H3b-qasper-hyb-d}) and loses on FreshStack (\R{H3b-fs-hyb-d}). On Qasper, S also beats the header-blind paragraph packer P by \R{SP-qasper-d} and length-matched windows FL by \R{SFL-qasper-d} (secondary), so on Qasper the gain appears to come from section boundaries, not from intact paragraphs or chunk length. On FreshStack both controls are within noise (\R{SP-fs}, \R{SFL-fs}); its gold spans are long sections, which overlapping windows serve as well. The section-path prefix itself has no detectable effect under the hybrid (S+T $-$ S: Qasper \R{STS-qasper}, FreshStack \R{STS-fs}; exploratory), so the structure gain comes from where chunks break, not from the prefix text.

\paragraph{Answer level.}
Qasper answer F1 is equivalent within $\pm$5 F1 for E $-$ S+T (\R{F1-H1-d}) and E $-$ CR (\R{F1-H2-d}). Every arm except F256t lies between \R{f1-qasper-S+T} and \R{f1-qasper-E-strong} F1 (Appendix Table~\ref{tab:answer}), and E changes recall on only \R{ax-f1-recchanged}\% of questions (exploratory), so with a 7B reader this equivalence is a weak check. On FreshStack the answer judge failed its negative control (\R{ctl-neg}\% against a limit of 5\%), so those results are exploratory.

\subsection{Stronger retrievers leave enrichment little to add}
\label{sec:why}

\begin{figure}[t]
\centering
\resizebox{\linewidth}{!}{
\begin{tikzpicture}[font=\scriptsize]
\draw[gray!60, dashed] (1.680,0) -- (1.680,4.960);
\draw (0.000,0) -- (0.000,-0.08) node[below] {$-$4};
\draw (0.840,0) -- (0.840,-0.08) node[below] {$-$2};
\draw (1.680,0) -- (1.680,-0.08) node[below] {0};
\draw (2.520,0) -- (2.520,-0.08) node[below] {+2};
\draw (3.360,0) -- (3.360,-0.08) node[below] {+4};
\draw (4.200,0) -- (4.200,-0.08) node[below] {+6};
\draw (5.040,0) -- (5.040,-0.08) node[below] {+8};
\draw (5.880,0) -- (5.880,-0.08) node[below] {+10};
\draw (0.000,0) -- (6.300,0);
\draw[gray!50] (0.000,3.100) -- (6.300,3.100);
\node at (3.150,-0.55) {E $-$ S+T (points; 95\% CI)};
\node[anchor=east] at (0.000,4.650) {BM25};
\draw (4.471,4.750) -- (6.236,4.750);
\node[fill=black, circle, inner sep=1.3pt] at (5.335,4.750) {};
\draw (2.050,4.550) -- (3.474,4.550);
\node[draw=black, fill=white, rectangle, inner sep=1.3pt] at (2.684,4.550) {};
\node[anchor=east] at (0.000,4.030) {Hybrid (primary)};
\draw (2.884,4.130) -- (5.091,4.130);
\node[fill=black, circle, inner sep=1.3pt] at (3.983,4.130) {};
\draw (1.906,3.930) -- (3.803,3.930);
\node[draw=black, fill=white, rectangle, inner sep=1.3pt] at (2.830,3.930) {};
\node[anchor=east] at (0.000,3.410) {BM25+qwen3};
\draw (2.525,3.510) -- (4.648,3.510);
\node[fill=black, circle, inner sep=1.3pt] at (3.594,3.510) {};
\draw (1.280,3.310) -- (2.890,3.310);
\node[draw=black, fill=white, rectangle, inner sep=1.3pt] at (2.116,3.310) {};
\node[anchor=east] at (0.000,2.790) {nomic};
\draw (0.980,2.890) -- (3.060,2.890);
\node[fill=black, circle, inner sep=1.3pt] at (2.041,2.890) {};
\draw (1.793,2.690) -- (3.796,2.690);
\node[draw=black, fill=white, rectangle, inner sep=1.3pt] at (2.782,2.690) {};
\node[anchor=east] at (0.000,2.170) {mxbai};
\draw (0.935,2.270) -- (3.071,2.270);
\node[fill=black, circle, inner sep=1.3pt] at (1.970,2.270) {};
\draw (0.730,2.070) -- (2.630,2.070);
\node[draw=black, fill=white, rectangle, inner sep=1.3pt] at (1.684,2.070) {};
\node[anchor=east] at (0.000,1.550) {jina-v3};
\draw (1.775,1.650) -- (4.005,1.650);
\node[fill=black, circle, inner sep=1.3pt] at (2.874,1.650) {};
\draw (0.158,1.450) -- (2.034,1.450);
\node[draw=black, fill=white, rectangle, inner sep=1.3pt] at (1.105,1.450) {};
\node[anchor=east] at (0.000,0.930) {qwen3};
\draw (1.274,1.030) -- (3.310,1.030);
\node[fill=black, circle, inner sep=1.3pt] at (2.317,1.030) {};
\draw (1.076,0.830) -- (2.538,0.830);
\node[draw=black, fill=white, rectangle, inner sep=1.3pt] at (1.790,0.830) {};
\node[anchor=east] at (0.000,0.310) {Hybrid+rerank};
\draw (0.981,0.410) -- (2.738,0.410);
\node[fill=black, circle, inner sep=1.3pt] at (1.846,0.410) {};
\draw (1.084,0.210) -- (2.935,0.210);
\node[draw=black, fill=white, rectangle, inner sep=1.3pt] at (2.011,0.210) {};
\node[fill=black, circle, inner sep=1.3pt] at (3.864,5.210) {};\node[anchor=west] at (3.990,5.210) {Qasper};
\node[draw=black, fill=white, rectangle, inner sep=1.3pt] at (5.292,5.210) {};\node[anchor=west] at (5.418,5.210) {FreshStack};
\end{tikzpicture}}
\caption{E $-$ S+T under eight retrievers, in points, with 95\% BCa cluster-bootstrap CIs. Filled circles: Qasper rec@512t; open squares: FreshStack prec@1024t. Rows above the divider rank with BM25, alone or fused; Hybrid+rerank re-scores the hybrid's top 30. Hybrid is the confirmatory H1 test. BM25, mxbai and qwen3 are secondary retrievers in the plan; BM25+qwen3, Hybrid+rerank and jina-v3 were added as secondary by Amendment~1; nomic alone is exploratory. CIs outside the primary row are unadjusted.}
\label{fig:retrievers}
\end{figure}

\paragraph{Under qwen3 and the reranker, enrichment is equivalent to the free prefix.}
With the 7B indexing LLM, the 95\% CIs of E $-$ S+T lie within the plan's margins on both datasets under qwen3 (Qasper \R{q3H1-qasper-d}, FreshStack \R{q3H1-fs-d}) and under the reranker (\R{rrH1-qasper-d}, \R{rrH1-fs-d}), as do those of E $-$ CR under the reranker (\R{rrH2-qasper-d}, \R{rrH2-fs-d}; Table~\ref{tab:confirm}, Figure~\ref{fig:retrievers}). On Qasper both CIs exclude a gain as large as the hybrid's (\R{H1-qasper-hyb-d}). Of qwen3, BM25+qwen3 and the reranker, only BM25+qwen3 keeps a Qasper gain (\R{hq3H1-qasper-d}); plain jina-v3 embeddings also show a small one (E \R{m-qasper-jina-E} against S+T \R{m-qasper-jina-S+T}; CI excludes 0 in Figure~\ref{fig:retrievers}).

\paragraph{Much of the hybrid gain repairs the fusion (exploratory).}
The primary hybrid, untuned reciprocal rank fusion fixed in the plan, scores below its own dense component on Qasper~\citep[cf.][]{bruch2023fusion}, and the loss falls mainly on the free prefix: hybrid enrichment scores \R{m-qasper-hybrid-E}, and the free prefix under nomic alone \R{m-qasper-nomic-S+T}. Nomic alone beats the hybrid by \R{ax-fr-qasper-STnomic-SThyb-d} points for S+T but by only \R{ax-fr-qasper-Enomic-Ehyb-d} for E (CI includes 0; exploratory), so fusing the weaker BM25 ranking pulls S+T down far more than E. On FreshStack nomic alone does not beat the hybrid.

\paragraph{Across retrievers (exploratory).}
The free prefix under qwen3 beats hybrid E by \R{ax-fr-qasper-STq3-Ehyb-d} points on Qasper, as much as enrichment adds under the hybrid, and by \R{ax-fr-fs-STq3-Ehyb-d} on FreshStack; under the reranker the differences are \R{ax-fr-qasper-STrr-Ehyb-d} and \R{ax-fr-fs-STrr-Ehyb-d}. With the same free prefix, qwen3 still beats nomic alone, which does not suffer the fusion loss, by \R{ax-fr-qasper-STq3-STnomic-d} points (FreshStack \R{ax-fr-fs-STq3-STnomic-d}). The 120B indexing LLM keeps a small edge at the top end: the best configuration of the grid (\CRstrong{} reranked; \Estrong{} ties) exceeds the best free one (S+T reranked) by \R{ax-fr-qasper-best-bestfree} on Qasper. Both were selected post hoc, and the CI does not account for that selection. On FreshStack the best configuration is LLM-free (P reranked, \R{ax-fr-fs-best-m}).

\paragraph{Why: enrichment acts as document expansion (exploratory).}
On Qasper the gain is mainly lexical. Qasper questions were written from a paper's title and abstract alone (\S\ref{sec:data}), so a question can use terms its evidence lacks. BM25 alone gains \R{H1-qasper-bm25} points, while nomic and mxbai alone each gain under a point and are equivalent. For \R{ax-br-qasper-share}\% of questions, the gold chunk's generated fields supply a question term that its section path and text lack. Under BM25 these bridged questions gain \R{ax-br-qasper-bm25-yes-d} points, against \R{ax-br-qasper-bm25-miss-d} for questions whose missing terms the fields do not supply, and they carry \R{ax-br-qasper-bm25-gainshare}\% of the BM25 gain and \R{ax-br-qasper-hyb-gainshare}\% of the hybrid gain. Questions with no missing term also gain (\R{ax-br-qasper-bm25-none-d}), so term reweighting contributes too. Overall question--evidence overlap does not predict the gain (lowest minus highest tercile, BM25: \R{ax-ovl-qasper-bm25-lmh}). On FreshStack the bridging contrast is within noise (BM25: \R{ax-br-fs-bm25-c}). Appendix~\ref{app:mechanism} breaks the gain down by retriever.

\subsection{Which add-ons earn their cost}
\label{sec:mechres}

None of the add-ons beats its LLM-free control (Appendix Table~\ref{tab:mech}). The dictionary raises key reuse (to \R{reuse-E-qasper}\% of Qasper chunks), yet retrieval is equivalent with and without it (H4d), presumably because the key itself is never indexed. Key merges beat random merges on Qasper only before adjustment (H4a \R{H4a-qasper-d}). Key-based merging changes Qasper recall by \R{H4c-qasper-d} points, a small loss within the margin, is equivalent on FreshStack, and does no better than merging the most similar neighbours, which needs no LLM call (H4b). A TopicGPT-style consolidation call lowers Qasper recall and raises FreshStack precision (H7); exploratory controls suggest that merging acts mainly through chunk size (Appendix~\ref{app:dial}). Related keys link \R{links-all-qasper}\% of Qasper chunks, so they barely select: expansion along them lowers Qasper recall (\R{H5a-qasper-d} points) at equal tokens and, before adjustment, does worse than adding the same number of adjacent chunks (\R{H5b-qasper}; \emph{n.d.}\ after Holm). Four calls per chunk, one per field, retrieve as well as the single call (H6, equivalent) at \R{pf-calls-ratio} times the calls~\citep[cf.][]{son2024mti}. Late chunking~\citep{gunther2024late}, an LLM-free way to give chunks document context, lowers Qasper recall (\R{lcS-qasper-d} points) against plain jina-v3 embeddings of the same chunks and has no detectable effect on FreshStack (\R{lcS-fs-d}; secondary).

\section{Where to Spend}
\label{sec:discussion}

\begin{table}[t]
\centering
\footnotesize
\setlength{\tabcolsep}{3pt}
\begin{tabularx}{\linewidth}{@{}l Y Y@{}}
\toprule
\textbf{Step} & \textbf{Do} & \textbf{Evidence} (Qasper; FreshStack) \\
\midrule
1 Retriever & stronger embedder over S+T; a reranker if the index cannot change & S+T (qwen3) $-$ E (hybrid): \R{ax-fr-qasper-STq3-Ehyb-d}; \R{ax-fr-fs-STq3-Ehyb-d}$^{x}$; E $-$ S+T (qwen3): \R{q3H1-qasper-d}; \R{q3H1-fs-d}$^{s}$ \\
2 Structure & header-led chunks; the section-path prefix is optional & S $-$ F512c: \R{H3a-qasper-hyb-d}; \R{H3a-fs-hyb-d}$^{c}$; S $-$ F256t, FreshStack: \R{H3b-fs-hyb-d}$^{c}$; S+T $-$ S: \R{STS-qasper-d}; \R{STS-fs-d}$^{x}$ \\
3 Enrichment & mainly for BM25-led stacks; one call, not CR & E $-$ S+T, BM25: \R{H1-qasper-bm25}; \R{H1-fs-bm25}$^{s}$; E $-$ CR equiv.$^{c}$ \\
4 Add-ons & skip key dictionary, key-based merging, related-key expansion, late chunking & Qasper: E+KX $-$ E \R{H5a-qasper-d}; late chunking $-$ S \R{lcS-qasper-d}$^{s}$ \\
\bottomrule
\end{tabularx}
\caption{Where to spend, in order. Differences in points of the primary metric. Tiers: $^{c}$confirmatory; $^{s}$pre-specified secondary; $^{x}$exploratory. Appendix Table~\ref{tab:guide} gives the full guide.}
\label{tab:guide-short}
\end{table}

\paragraph{An order of spending.}
For corpora and models like ours, Table~\ref{tab:guide-short} orders the choices by what they returned; Figure~\ref{fig:frontier} shows why. Spend first on the retriever: a stronger embedder if the corpus can be re-embedded, a reranker if it cannot. The two cost differently. A new embedder is a one-time indexing cost with no LLM call; a reranker adds a per-query cost (30 candidates per query), so it is the cheaper option where query volume is low relative to corpus size. We did not measure query-time cost. A stack that fuses BM25 and dense rankings without tuning should check that the fusion beats its dense part. Next, use header-led chunks, which cost nothing; the section-path prefix had no detectable effect, and where gold evidence spans whole sections, overlapping token windows did as well or better. Enrich where BM25 drives the ranking, alone or in an untuned fusion. As in doc2query~\citep{nogueira2019doc2query}, the fields add query terms the chunk lacks, and as \citet{weller2024expansions} report for expansions in general, they help weak retrievers; with a strong retriever and the 7B indexing LLM we find equivalence rather than a loss. A team that does enrich needs one call for all fields, not CR's whole-document prompt. Key dictionaries, key-based merging, related-key expansion and late chunking did not earn their cost here, in line with \citet{caspari2026complex}, \citet{qu2025semantic} and \citet{zhou2026beyond}.

\paragraph{When enrichment may still pay.}
The expansion gain comes largely from fields that supply a question term the chunk lacks (\S\ref{sec:why}), which Qasper's question collection favours: its questions were written without the paper's body. On FreshStack, whose questions are Stack Overflow posts about the documented libraries, bridging was within noise. Corpora whose users query in the documents' own vocabulary should expect less from enrichment; corpora searched with lexical retrieval and outsider vocabulary should expect more. Larger indexing LLMs may also keep a small edge under strong retrievers: the best 120B-indexed configuration leads the best free one by \R{ax-fr-qasper-best-bestfree-d} on Qasper (exploratory, selected post hoc), which no planned test covers.

\paragraph{Cheap controls change the conclusions.}
Every control that changed a conclusion here needs no LLM call: the free prefix, count-matched random and similarity merges, equal-token neighbour expansion, and a second retriever. Without a second retriever, the \R{H1-qasper-hyb-d} points would read as the return on an LLM call. Planned margins and sample sizes mattered as much: the pilot was inconclusive on every hybrid enrichment comparison.

\section{Conclusion}
\label{sec:conclusion}

Under a plan made public before any index was built, one LLM enrichment call per chunk improved Qasper retrieval under a default hybrid retriever (on FreshStack the gain did not survive Holm adjustment) and matched contextual retrieval at a fraction of its input tokens; on Qasper the gain behaved as document expansion (exploratory). Under qwen3-embedding or a reranker, the two strongest retrievers we tested (pre-specified secondary), 7B enrichment was equivalent to the free prefix. Header-led chunks beat 512-character windows but lost to 256-token windows on FreshStack; no add-on beat its LLM-free control. For header-structured corpora: check that the hybrid fusion beats its dense part, upgrade the retriever, chunk at headers, and enrich mainly for lexical retrieval (Table~\ref{tab:guide-short}).

\section*{Limitations}

\paragraph{Models and prompts.}
The indexed text comes from two indexing LLMs, with one prompt and one generation run per configuration, and the answers come from one fixed 7B reader. The plan's robustness runs (repeat generations, a paraphrased prompt) were not run, so the CIs cover the sampling of questions, not variation across runs or prompts. Other prompts or larger models may change the size of the enrichment effects.

\paragraph{Data and retrieval settings.}
We test two English benchmarks. Qasper papers are converted from JSON and searched one paper at a time; FreshStack's gold labels are LLM judgments that often span whole sections, and its answer-level judge failed its negative control. Both benchmarks mostly need one passage or one section, so methods that gather evidence scattered across a document are under-tested. Retrieval runs within single papers or within capped documentation corpora, so effects at the scale of large multi-document corpora, where contextual text may also disambiguate between documents~\citep{reuter2025sac}, are untested. All retrievers are small, the primary hybrid uses untuned rank fusion by design, the reranker was run over the hybrid only, and we tested no learned sparse, multi-vector or doc2query baseline. Our cost comparison counts indexing-time LLM tokens and API spend only; the query-time cost of rerankers and larger embedders was not measured.

\paragraph{Design.}
The plan was public before any index was built but was not registered with an external registry, and it was written after an exploratory pilot had been viewed. The equivalence margins were chosen for statistical power, not as a smallest effect of practical interest; a $\pm$5 margin is close to the H1 effect itself, so equivalence excludes gains of that size, not small ones. The add-on tests were specified after the main results had been viewed, so we report them as secondary. Analyses not named in the plan or its amendments, including every contrast across retrievers, are exploratory and need confirmation.

\paragraph{Implementation scope.}
All arms use one implementation of header-led chunking and of enrichment, run with the harness's LLM settings (model, temperature, seed, output limits) and with key-based merging only in the merge arms; chunking uses the package defaults. The add-ons, including the rolling dictionary and its size and eviction parameters, were tested in one design and with the 7B indexing LLM only. Other designs of label reuse, merging or linking may behave differently.

\section*{Ethics Statement}

\ifreview\else The authors developed the open-source system under test, so they evaluate their own system.\fi{} We limited the resulting bias in three ways. The hypotheses, metrics, margins and decision rules were fixed in a plan made public before any index was built, later pre-specified tests were added only through time-stamped amendments labelled secondary, and all post-hoc analyses are labelled exploratory. Every add-on is compared with matched controls that need no call to our system. A separate re-computation from the per-question outputs checked every confirmatory and add-on difference and interval, and its corrections are listed in Appendix~\ref{app:plan}. Most of our findings about the system's own add-ons are negative, and we report them as such. We use two public research datasets, Qasper and FreshStack; FreshStack's queries come from public Stack Overflow posts and its corpora from open-source documentation repositories. Qasper is released under CC BY 4.0 and FreshStack under CC BY-SA 4.0, the licence of Stack Overflow posts; the corpus repositories are under MIT, BSD, Apache-2.0, CC BY 4.0, GPL-3.0 and AGPL-3.0 licences. We use them for research evaluation only; released files that contain corpus text keep the source licences, including ShareAlike. No new human-subject data were collected. API calls were capped at US\$5.50 and logged (actual spend US\$\R{spend}), with provider data collection disabled. Local generation ran on one consumer laptop. A null or equivalence result should not be read as applying beyond the tested models and corpora. AI assistants were used for coding and for editing the text.

\bibliography{references}

\appendix

\section{Practitioner Guide}
\label{app:guide}

Table~\ref{tab:guide} maps common situations to the configuration our results support, with the evidence for each row. Figure~\ref{fig:frontier} (\S\ref{sec:intro}) shows the underlying trade-off between indexing spend and retriever choice; Table~\ref{tab:guide-short} is the short version. Rows 1--2 and the untuned-RRF row rest on exploratory cross-retriever contrasts, and E+AX $-$ E is exploratory; the other rows rest on confirmatory, add-on or secondary-retriever tests. The guide is limited to the tested models, retrievers and corpora (see Limitations).

\begin{table*}[!htbp]
\centering
\small
\setlength{\tabcolsep}{4pt}
\begin{tabularx}{\textwidth}{@{}Y Y l@{}}
\toprule
\textbf{Situation} & \textbf{Use} & \textbf{Evidence} (Qasper; FreshStack) \\
\midrule
Can re-embed the corpus & qwen3-class embedder over S+T; skip enrichment & S+T (qwen3) $-$ E (hybrid): \R{ax-fr-qasper-STq3-Ehyb-d}; \R{ax-fr-fs-STq3-Ehyb-d} \\
Cannot re-index & reranker over the existing hybrid & S+T (rerank) $-$ E (hybrid): \R{ax-fr-qasper-STrr-Ehyb-d}; \R{ax-fr-fs-STrr-Ehyb-d} \\
Strong retriever already in place & skip 7B enrichment; 120B indexing (E$_s$ or CR$_s$) kept \R{ax-fr-qasper-best-bestfree-d} on Qasper at the top end (exploratory, best of the grid) & E $-$ S+T, qwen3: \R{q3H1-qasper-d}; \R{q3H1-fs-d}; rerank: \R{rrH1-qasper-d}; \R{rrH1-fs-d} \\
Keyword-only stack (no vectors, no reranker) & enrichment (document expansion) & E $-$ S+T, BM25: \R{H1-qasper-bm25}; \R{H1-fs-bm25} \\
Paying for an indexing call & enrichment, not CR & H2 equiv.; CR sends $\times$\R{crratio-local-qasper}--\R{crratio-local-fs} input tokens \\
One call or one per field & one call & H6 equiv.; per-field calls cost $\times$\R{pf-calls-ratio} calls \\
Chunk boundaries & header-led chunks for paragraph-scale evidence; on section-scale gold, 256-token overlapping windows did as well or better & H3a (S $-$ F512c): \R{H3a-qasper-hyb-d}; \R{H3a-fs-hyb-d}; H3b FS: \R{H3b-fs-hyb-d} \\
Within-document search & avoid late chunking & late chunking $-$ S: \R{lcS-qasper-d}; \R{lcS-fs-d} \\
Merging chunks & no merging by default; if merging, similarity merges showed no detectable difference from key-based merges & Qasper H4c (M $-$ E): \R{H4c-qasper-d}; H4b (M $-$ SM): \R{H4b-qasper-d}; \R{H4b-fs-d} \\
Untuned RRF hybrid & check it beats its dense part before paying for enrichment & nomic $-$ hybrid, S+T: \R{ax-fr-qasper-STnomic-SThyb-d}; E: \R{ax-fr-qasper-Enomic-Ehyb-d} \\
Fixed-budget expansion & avoid for paragraph-scale evidence & Qasper: E+KX $-$ E: \R{H5a-qasper-d}; E+AX $-$ E: \R{EAX-E-qasper-d} \\
\bottomrule
\end{tabularx}
\caption{Practitioner guide. Differences in points of the primary metric; rows 1--2, the untuned-RRF row and E+AX $-$ E are exploratory (\S\ref{sec:why}); the others are confirmatory, add-on or secondary-retriever results. The guide reflects the tested models and corpora.}
\label{tab:guide}
\end{table*}

\section{Answer-Level Outcomes}
\label{app:answer}

Table~\ref{tab:answer} gives answer-level outcomes for the main arms under the primary hybrid retriever (\S\ref{sec:confirm}).

\paragraph{Answers track the retrieved evidence (descriptive).}
Answer-level equivalence averages over changes that cancel. E changes recall on only \R{ax-f1-recchanged}\% of Qasper questions. Where it retrieves more evidence (\R{ax-f1-recup-n} questions), F1 changes by \R{ax-f1-recup}; where it retrieves less (\R{ax-f1-recdown-n}), by \R{ax-f1-recdown}; where it is unchanged (\R{ax-f1-recsame-n}), by \R{ax-f1-recsame}. The groups are defined by an outcome, so these are descriptive conditional means, not effects of enrichment; they show only that the reader is sensitive to its evidence.

\begin{table}[!htbp]
\centering
\footnotesize
\setlength{\tabcolsep}{3pt}
\resizebox{\linewidth}{!}{%
\begin{tabular}{@{}lcc@{}}
\toprule
\textbf{Arm} & \textbf{Qasper F1} & \textbf{FreshStack nugget rec.}$^{\dagger}$ \\
\midrule
F512c & \R{f1-qasper-F512c} & \R{nug-F512c} \\
F256t & \R{f1-qasper-F256t} & \R{nug-F256t} \\
S & \R{f1-qasper-S} & \R{nug-S} \\
S+T & \R{f1-qasper-S+T} & \R{nug-S+T} \\
E & \R{f1-qasper-E} & \R{nug-E} \\
CR & \R{f1-qasper-CR} & \R{nug-CR} \\
\Estrong{} & \R{f1-qasper-E-strong} & \R{nug-E-strong} \\
\midrule
E $-$ S+T & \R{F1-H1}, \R{F1-H1-v} & \R{nug-E-SpT} \\
E $-$ CR & \R{F1-H2}, \R{F1-H2-v} & \R{nug-E-CR} \\
\bottomrule
\end{tabular}}
\caption{Answer-level outcomes under hybrid retrieval, in points. The reader is \texttt{qwen2.5:7b} at temperature 0, given 512 (Qasper) or 1,024 (FreshStack) tokens of retrieved source text. Qasper differences: 90\% CI with the key-secondary verdict (margin $\pm$5). $^{\dagger}$Exploratory: strict nugget recall judged by \texttt{llama-3.3-70b}, whose negative control failed (\R{ctl-neg}\% $>$ 5\%); 95\% CIs.}
\label{tab:answer}
\end{table}

\section{Secondary Retrievers and Indexing-LLM Scaling}
\label{app:secondary}

All embedders use their documented prefixes: nomic's \texttt{search\_query:}/\texttt{search\_document:}, mxbai's query instruction, a retrieval task instruction for qwen3-embedding queries, and jina-v3's \texttt{retrieval.query}/\texttt{retrieval.passage} adapters. Table~\ref{tab:secondary} reports the alignment controls, the secondary retrievers and exploratory cross-retriever contrasts; Figure~\ref{fig:retrievers} plots E $-$ S+T under every retriever. The reranker is Qwen3-Reranker-0.6B over the hybrid (BM25+nomic) top-30; it was not run over any other retriever. We run its sentence-transformers CrossEncoder port with the Qwen3 instruction template; it scores each candidate's indexed text (generated fields, section path or CR context, plus the chunk) with the query, truncated to 2,048 tokens. Under the reranker, the 95\% CIs of E $-$ S+T and E $-$ CR lie within the plan's margins on both datasets; under qwen3-embedding, those of E $-$ S+T do. Reranking lifts the weaker arms most: Qasper S+T rises from \R{m-qasper-S+T} to \R{m-qasper-hybrid+rerank-S+T} and E from \R{m-qasper-E} to \R{m-qasper-hybrid+rerank-E}. On Qasper the gain is largest where BM25 contributes to the ranking: BM25 alone (\R{H1-qasper-bm25}), the primary hybrid (\R{H1-qasper-hyb-d}) and the BM25+qwen3 hybrid (\R{hq3H1-qasper-d}). This fits the lexical reading in \S\ref{sec:why}. On FreshStack the gain is detectable under BM25 (\R{H1-fs-bm25}), nomic alone (\R{H1-fs-nomic-d}) and, before Holm adjustment, the hybrid (\R{H1-fs-hyb-d}) but not under qwen3, the BM25+qwen3 hybrid or the reranker.

Late chunking (jina-embeddings-v3, 8,192-token windows with 2,048 tokens of overlap) lowers Qasper recall against plain embeddings of the same header chunks (LC-S $-$ S: \R{lcS-qasper}) and has no detectable effect on FreshStack (\R{lcS-fs}). Every chunk span was located in its source document, none fell back to a plain embedding, and none was split across windows. With whole-document context, the mean cosine between chunks of one document rises from \R{lcsim-qasper-plain} to \R{lcsim-qasper-late} on Qasper (\R{lcsim-qasper-n} papers) and from \R{lcsim-fs-plain} to \R{lcsim-fs-late} on FreshStack (\R{lcsim-fs-n} documents). This rise does not explain the Qasper loss (exploratory): the per-paper loss is unrelated to the rise in cosine (Spearman \R{ax-lc-rho-loss-dcos}) and to the paper's chunk count (\R{ax-lc-rho-loss-nchunks}), and the \R{ax-lc-win-one-n} papers that fit in one window change by \R{ax-lc-win-one-d} points, against \R{ax-lc-win-multi-d} for the \R{ax-lc-win-multi-n} papers that need several (contrast \R{ax-lc-win-c}). Enrichment beats late chunking on Qasper (E $-$ LC-S: \R{lcE-qasper}).

\begin{table}[!htbp]
\centering
\footnotesize
\setlength{\tabcolsep}{3pt}
\resizebox{\linewidth}{!}{%
\begin{tabular}{@{}lcc@{}}
\toprule
\textbf{A $-$ B (retriever)} & \textbf{Qasper} & \textbf{FreshStack} \\
\midrule
S $-$ P (hybrid) & \R{SP-qasper} & \R{SP-fs} \\
S $-$ FL (hybrid) & \R{SFL-qasper} & \R{SFL-fs} \\
S+T $-$ S (hybrid) & \R{STS-qasper} & \R{STS-fs} \\
\Estrong{} $-$ E (hybrid) & \R{EsE-qasper} & \R{EsE-fs} \\
E $-$ S+T (hybrid+rerank) & \R{rrH1-qasper} & \R{rrH1-fs} \\
E $-$ CR (hybrid+rerank) & \R{rrH2-qasper} & \R{rrH2-fs} \\
E $-$ S+T (qwen3-emb.) & \R{q3H1-qasper} & \R{q3H1-fs} \\
E $-$ S+T (BM25+qwen3) & \R{hq3H1-qasper} & \R{hq3H1-fs} \\
LC-S $-$ S (jina-v3) & \R{lcS-qasper} & \R{lcS-fs} \\
E $-$ LC-S (jina-v3) & \R{lcE-qasper} & \R{lcE-fs} \\
\midrule
\multicolumn{3}{@{}l}{\emph{Exploratory cross-retriever contrasts}} \\
S+T (qwen3) $-$ S+T (rerank) & \R{ax-fr-qasper-STq3-STrr} & \R{ax-fr-fs-STq3-STrr} \\
S+T (qwen3) $-$ E (hybrid) & \R{ax-fr-qasper-STq3-Ehyb} & \R{ax-fr-fs-STq3-Ehyb} \\
S+T (rerank) $-$ E (hybrid) & \R{ax-fr-qasper-STrr-Ehyb} & \R{ax-fr-fs-STrr-Ehyb} \\
S+T (nomic) $-$ S+T (hybrid) & \R{ax-fr-qasper-STnomic-SThyb} & \R{ax-fr-fs-STnomic-SThyb} \\
E (nomic) $-$ E (hybrid) & \R{ax-fr-qasper-Enomic-Ehyb} & \R{ax-fr-fs-Enomic-Ehyb} \\
Best $-$ best free config. & \R{ax-fr-qasper-best-bestfree} & -- (best is free) \\
\bottomrule
\end{tabular}}
\caption{Secondary comparisons in points, with unadjusted 95\% BCa CIs. rerank: Qwen3-Reranker-0.6B over the hybrid top-30. LC-S: late chunking over header-led chunks. Best configurations over all arms and retrievers: Qasper \CRstrong{} reranked (\R{ax-fr-qasper-best-m}) against S+T reranked (\R{ax-fr-qasper-bestfree-m}); on FreshStack the best configuration, P reranked (\R{ax-fr-fs-best-m}), is free. The best configurations were selected post hoc; their CI does not account for the selection. Full grids are in Appendix~\ref{app:grids}.}
\label{tab:secondary}
\end{table}

\paragraph{Indexing-LLM scaling (exploratory).}
On \R{scale-n} of the 30 sampled Qasper papers (the API quota ran out; Appendix~\ref{app:plan}), E built with Qwen3-235B-A22B scored \R{scale-E-qwen235b} against \R{scale-E} for the 7B model and \R{scale-E-strong} for \texttt{gpt-oss-120b}. The difference from the 7B model (\R{scale-E-qwen235b-E}) has a CI that includes 0, so the sample cannot tell whether a larger indexing LLM helps.

\section{Exploratory Analyses of the Enrichment Gain}
\label{app:mechanism}

\begin{figure*}[!htbp]
\centering
\begin{tikzpicture}[font=\scriptsize]
\draw[gray!25] (0,0.000) -- (9.00,0.000);\node[left] at (0,0.000) {$-$8};
\draw[gray!25] (0,0.680) -- (9.00,0.680);\node[left] at (0,0.680) {0};
\draw[gray!25] (0,1.360) -- (9.00,1.360);\node[left] at (0,1.360) {+8};
\draw[gray!25] (0,2.040) -- (9.00,2.040);\node[left] at (0,2.040) {+16};
\draw[gray!25] (0,2.720) -- (9.00,2.720);\node[left] at (0,2.720) {+24};
\draw[gray!25] (0,3.400) -- (9.00,3.400);\node[left] at (0,3.400) {+32};
\draw (0,0.680) -- (9.00,0.680);
\node[below] at (0.75,0) {BM25};
\filldraw[fill=black, draw=black, line width=0.3pt] (0.300,0.680) rectangle (0.600,2.754);
\draw[line width=0.4pt] (0.450,2.308) -- (0.450,3.243);
\filldraw[fill=white, draw=black, line width=0.3pt] (0.600,0.680) rectangle (0.900,0.950);
\draw[line width=0.4pt] (0.750,0.753) -- (0.750,1.153);
\filldraw[fill=gray!45, draw=black, line width=0.3pt] (0.900,0.680) rectangle (1.200,1.835);
\draw[line width=0.4pt] (1.050,1.305) -- (1.050,2.478);
\node[below] at (2.25,0) {Hybrid};
\filldraw[fill=black, draw=black, line width=0.3pt] (1.800,0.680) rectangle (2.100,2.229);
\draw[line width=0.4pt] (1.950,1.713) -- (1.950,2.753);
\filldraw[fill=white, draw=black, line width=0.3pt] (2.100,0.680) rectangle (2.400,0.798);
\draw[line width=0.4pt] (2.250,0.538) -- (2.250,1.056);
\filldraw[fill=gray!45, draw=black, line width=0.3pt] (2.400,0.680) rectangle (2.700,1.304);
\draw[line width=0.4pt] (2.550,0.781) -- (2.550,1.850);
\node[below] at (3.75,0) {BM25+qwen3};
\filldraw[fill=black, draw=black, line width=0.3pt] (3.300,0.680) rectangle (3.600,2.049);
\draw[line width=0.4pt] (3.450,1.624) -- (3.450,2.531);
\filldraw[fill=white, draw=black, line width=0.3pt] (3.600,0.680) rectangle (3.900,0.718);
\draw[line width=0.4pt] (3.750,0.471) -- (3.750,0.974);
\filldraw[fill=gray!45, draw=black, line width=0.3pt] (3.900,0.680) rectangle (4.200,1.394);
\draw[line width=0.4pt] (4.050,0.840) -- (4.050,1.972);
\node[below] at (5.25,0) {nomic};
\filldraw[fill=black, draw=black, line width=0.3pt] (4.800,0.680) rectangle (5.100,1.257);
\draw[line width=0.4pt] (4.950,0.791) -- (4.950,1.710);
\filldraw[fill=white, draw=black, line width=0.3pt] (5.100,0.680) rectangle (5.400,0.618);
\draw[line width=0.4pt] (5.250,0.342) -- (5.250,0.889);
\filldraw[fill=gray!45, draw=black, line width=0.3pt] (5.400,0.680) rectangle (5.700,0.678);
\draw[line width=0.4pt] (5.550,0.134) -- (5.550,1.211);
\node[below] at (6.75,0) {qwen3};
\filldraw[fill=black, draw=black, line width=0.3pt] (6.300,0.680) rectangle (6.600,1.101);
\draw[line width=0.4pt] (6.450,0.726) -- (6.450,1.501);
\filldraw[fill=white, draw=black, line width=0.3pt] (6.600,0.680) rectangle (6.900,0.735);
\draw[line width=0.4pt] (6.750,0.455) -- (6.750,1.017);
\filldraw[fill=gray!45, draw=black, line width=0.3pt] (6.900,0.680) rectangle (7.200,0.741);
\draw[line width=0.4pt] (7.050,0.215) -- (7.050,1.286);
\node[below] at (8.25,0) {Hyb.+rerank};
\filldraw[fill=black, draw=black, line width=0.3pt] (7.800,0.680) rectangle (8.100,0.985);
\draw[line width=0.4pt] (7.950,0.578) -- (7.950,1.381);
\filldraw[fill=white, draw=black, line width=0.3pt] (8.100,0.680) rectangle (8.400,0.629);
\draw[line width=0.4pt] (8.250,0.408) -- (8.250,0.859);
\filldraw[fill=gray!45, draw=black, line width=0.3pt] (8.400,0.680) rectangle (8.700,0.737);
\draw[line width=0.4pt] (8.550,0.475) -- (8.550,1.028);
\draw (0,0) rectangle (9.00,3.40);
\node[rotate=90] at (-0.75,1.70) {E $-$ S+T, Qasper (points)};
\filldraw[fill=black, draw=black] (0.00,3.58) rectangle (0.18,3.76);\node[anchor=west] at (0.22,3.67) {missing term supplied};
\filldraw[fill=white, draw=black] (3.20,3.58) rectangle (3.38,3.76);\node[anchor=west] at (3.42,3.67) {missing term not supplied};
\filldraw[fill=gray!45, draw=black] (6.90,3.58) rectangle (7.08,3.76);\node[anchor=west] at (7.12,3.67) {no missing term};
\end{tikzpicture}
\caption{Qasper E $-$ S+T (points, 95\% cluster-bootstrap CIs, exploratory) for three question groups: the generated fields of the gold chunk supply a question term missing from its section path and text (\R{ax-br-qasper-share}\% of questions); a term is missing but not supplied; no term is missing. The supplied-term group gains most under every retriever, most of all where BM25 enters the ranking.}
\label{fig:mechanism}
\end{figure*}

All analyses in this appendix and in \S\ref{sec:why}, except the tests under the secondary retrievers, are post hoc. They were written after every confirmatory, secondary and add-on result had been viewed, use only existing per-question outputs and chunk caches (no new build, retrieval run or model call), and are released as \texttt{analysis\_v4.py}. Statistics reuse the harness: paired differences, cluster bootstrap (papers on Qasper, questions within domains on FreshStack), 10,000 resamples, unadjusted 95\% BCa CIs. Contrasts between question groups resample clusters jointly. Every analysis that was run is reported, including nulls. Figure~\ref{fig:mechanism} summarises the document-expansion analysis.

\paragraph{Definitions.}
Question terms are the question's word types after removing English stop words and one-character tokens. The \emph{gold chunks} of a question are the header-led (S) chunks that cover at least one of its evidence units under the matching rule of \S\ref{sec:metrics}. \emph{Overlap} is the share of question terms that occur in the gold evidence text. A question is \emph{bridged} if some question term is absent from its gold chunks' S+T text (section path and chunk text) but present in their E rendering, that is, supplied by the generated fields. \emph{Targeting} is the number of question terms in the gold chunks' generated fields minus the mean number in the generated fields of the other chunks of the same paper or domain.

\paragraph{Vocabulary overlap.}
With questions split into overlap terciles, the lowest minus highest tercile difference in E $-$ S+T is \R{ax-ovl-qasper-bm25-lmh} under BM25, \R{ax-ovl-qasper-hyb-lmh} under the hybrid and \R{ax-ovl-qasper-nomic-lmh} under nomic on Qasper, and \R{ax-ovl-fs-bm25-lmh} under BM25 on FreshStack. Vocabulary mismatch alone does not moderate the gain.

\paragraph{Bridging.}
On Qasper, \R{ax-br-qasper-n} questions (\R{ax-br-qasper-share}\%) are bridged; \R{ax-br-qasper-nmiss} have a missing term that the fields do not supply and \R{ax-br-qasper-nnone} have no missing term. E $-$ S+T by group:

\begin{center}
\scriptsize
\setlength{\tabcolsep}{3pt}
\begin{tabular}{@{}lccc@{}}
\toprule
\textbf{Retriever} & \textbf{Bridged} & \textbf{Missing, not bridged} & \textbf{Share of gain} \\
\midrule
BM25 & \R{ax-br-qasper-bm25-yes} & \R{ax-br-qasper-bm25-miss} & \R{ax-br-qasper-bm25-gainshare}\% \\
Hybrid & \R{ax-br-qasper-hyb-yes} & \R{ax-br-qasper-hyb-miss} & \R{ax-br-qasper-hyb-gainshare}\% \\
BM25+qwen3 & \R{ax-br-qasper-hq3-yes} & \R{ax-br-qasper-hq3-miss} & \R{ax-br-qasper-hq3-gainshare}\% \\
qwen3 & \R{ax-br-qasper-q3-yes} & \R{ax-br-qasper-q3-miss} & \R{ax-br-qasper-q3-gainshare}\% \\
Rerank & \R{ax-br-qasper-rr-yes} & \R{ax-br-qasper-rr-miss} & -- \\
\bottomrule
\end{tabular}
\end{center}

\noindent The bridged minus not-bridged contrast is \R{ax-br-qasper-bm25-c} under BM25 and \R{ax-br-qasper-hyb-c} under the hybrid. Shares of the gain are omitted where the average gain is near zero, because the share is then unstable. Questions with no missing term also gain under BM25 (\R{ax-br-qasper-bm25-none}), so bridging is not the only route. On FreshStack \R{ax-br-fs-share}\% of questions are bridged and the contrast is null (BM25 \R{ax-br-fs-bm25-c}, hybrid \R{ax-br-fs-hyb-c}).

\paragraph{Targeting.}
On Qasper the gold chunks' fields contain \R{ax-tg-qasper-gold} question terms on average, against \R{ax-tg-qasper-other} for other chunks. E $-$ S+T by targeting tercile (bottom / middle / top) is \R{ax-tg-qasper-bm25-low-d} / \R{ax-tg-qasper-bm25-mid-d} / \R{ax-tg-qasper-bm25-high-d} under BM25, \R{ax-tg-qasper-hyb-low-d} / \R{ax-tg-qasper-hyb-mid-d} / \R{ax-tg-qasper-hyb-high-d} under the hybrid, \R{ax-tg-qasper-q3-low-d} / \R{ax-tg-qasper-q3-mid-d} / \R{ax-tg-qasper-q3-high-d} under qwen3 and \R{ax-tg-qasper-rr-low-d} / \R{ax-tg-qasper-rr-mid-d} / \R{ax-tg-qasper-rr-high-d} under the reranker. The top minus bottom contrast excludes 0 under every retriever (BM25 \R{ax-tg-qasper-bm25-c}, hybrid \R{ax-tg-qasper-hyb-c}, nomic \R{ax-tg-qasper-nomic-c}, mxbai \R{ax-tg-qasper-mxbai-c}, qwen3 \R{ax-tg-qasper-q3-c}, BM25+qwen3 \R{ax-tg-qasper-hq3-c}, rerank \R{ax-tg-qasper-rr-c}). On FreshStack no contrast excludes 0 (BM25 \R{ax-tg-fs-bm25-c}, hybrid \R{ax-tg-fs-hyb-c}).

\paragraph{Answer type, budgets and win rates.}
On Qasper, yes/no questions gain no recall under the hybrid (\R{ax-het-qasper-answertype-yesno-hyb}), while extractive questions gain \R{ax-het-qasper-answertype-extractive-hyb}. Under the hybrid, E beats S+T on \R{ax-wtl-qasper-hyb-win}\% of Qasper questions, ties on \R{ax-wtl-qasper-hyb-tie}\% and loses on \R{ax-wtl-qasper-hyb-loss}\%. On FreshStack the gain appears only at precision budgets of at least 1,024 tokens (prec@512t \R{ax-bud-fs-E-SpT-hyb-prec@512t}; prec@2048t \R{ax-bud-fs-E-SpT-hyb-prec@2048t}).

\section{Add-on Tests}
\label{app:components}

Add-on arms reuse the cached E enrichments, the hybrid retriever, the primary metrics and margins, and the rules of \S\ref{sec:stats}. Table~\ref{tab:mech} gives every pre-specified add-on test.

\paragraph{Merging (H4a--c).} M applies key-based merging to E. Two controls make the same number of merges in every document: RM merges random adjacent pairs (5 seeds, averaged per question) and SM greedily merges the adjacent pair whose chunk embeddings (nomic, mean over members) are most similar. Both respect the 3,000-character cap and, unlike key-based merging, merge only adjacent chunks. H4a tests M $-$ RM and H4b M $-$ SM (two-sided); H4c tests M $-$ E (TOST).

\paragraph{Dictionary (H4d, H7).} E$-$K repeats E with the key list replaced by ``(not provided)''. H4d tests E $-$ E$-$K (TOST). The H4 tests on two datasets form one family of eight. H7 compares the in-call dictionary with the TopicGPT pattern~\citep{pham2024topicgpt}: one call per document maps each E$-$K key to a canonical key from the same list, and M$-$TP merges on the consolidated keys. H7 tests M $-$ M$-$TP (two-sided; family of two).

\paragraph{Related keys (H5).} E+KX inserts, after each retrieved chunk, the not-yet-included chunks of its document that share its key or are linked to it through related keys. The control E+AX inserts approximately the same number of chunks, taken from the retrieved chunk's neighbours in the document. The token budget cuts the list as usual, so arms are compared at equal tokens. H5a tests E+KX $-$ E and H5b E+KX $-$ E+AX (two-sided; family of four). Section expansion (E+SX, parent-section retrieval) is exploratory.

\paragraph{Single call (H6).} PF makes one call per indexed field, with that field's instruction copied from the enrichment prompt and the same context, on a seeded 100 of the Qasper papers (317 questions). H6 tests E $-$ PF (TOST, $\pm$5).

\paragraph{Results.} The dictionary raises key reuse from \R{reuse-E-noRK-qasper}\% to \R{reuse-E-qasper}\% of Qasper chunks and from \R{reuse-E-noRK-fs}\% to \R{reuse-E-fs}\% on FreshStack. Key merges beat random merges on Qasper only before adjustment. H5b on Qasper is \R{H5b-qasper}: worse than the adjacency control before adjustment, \emph{n.d.}\ after Holm.

\begin{table}[!htbp]
\centering
\footnotesize
\setlength{\tabcolsep}{4pt}
\begin{tabular}{@{}llcc@{}}
\toprule
\textbf{H} & \textbf{A $-$ B} & \textbf{Diff.} [CI] & \textbf{Verdict} \\
\midrule
\multicolumn{4}{@{}l}{\emph{Qasper} (rec@512t)} \\
H4a & M $-$ RM & \R{H4a-qasper} & \R{H4a-qasper-v} \\
H4b & M $-$ SM & \R{H4b-qasper} & \R{H4b-qasper-v} \\
H4c & M $-$ E & \R{H4c-qasper} & \R{H4c-qasper-v} \\
H4d & E $-$ E$-$K & \R{H4d-qasper} & \R{H4d-qasper-v} \\
H7 & M $-$ M$-$TP & \R{H7-qasper} & \R{H7-qasper-v} \\
H5a & E+KX $-$ E & \R{H5a-qasper} & \R{H5a-qasper-v} \\
H5b & E+KX $-$ E+AX & \R{H5b-qasper} & \R{H5b-qasper-v} \\
H6 & E $-$ PF & \R{H6} & \R{H6-v} \\
\multicolumn{4}{@{}l}{\emph{FreshStack} (prec@1024t)} \\
H4a & M $-$ RM & \R{H4a-fs} & \R{H4a-fs-v} \\
H4b & M $-$ SM & \R{H4b-fs} & \R{H4b-fs-v} \\
H4c & M $-$ E & \R{H4c-fs} & \R{H4c-fs-v} \\
H4d & E $-$ E$-$K & \R{H4d-fs} & \R{H4d-fs-v} \\
H7 & M $-$ M$-$TP & \R{H7-fs} & \R{H7-fs-v} \\
H5a & E+KX $-$ E & \R{H5a-fs} & \R{H5a-fs-v} \\
H5b & E+KX $-$ E+AX & \R{H5b-fs} & \R{H5b-fs-v} \\
\bottomrule
\end{tabular}
\caption{Add-on tests (Amendments 2--4; pre-specified secondary), hybrid retriever, paired differences in points. M: key-based merging; RM, SM: count-matched random and similarity merges; E$-$K: no key dictionary; M$-$TP: key-based merging on keys consolidated after the fact; KX, AX: related-key and adjacent expansion; PF: one call per field (100 Qasper papers). CIs: 90\% for TOST rows (H4c, H4d, H6), 95\% otherwise. Holm families: H4 (8 tests), H7 (2), H5 (4). n.d.: no difference detected.}
\label{tab:mech}
\end{table}

\section{Merging and Chunk Size (Exploratory)}
\label{app:dial}

\begin{table}[!htbp]
\centering
\footnotesize
\setlength{\tabcolsep}{4pt}
\begin{tabular}{@{}llrrc@{}}
\toprule
 & \textbf{Keys} & \textbf{Reuse} & \textbf{Merges} & $\Delta$ [95\% CI] \\
\midrule
\multirow{3}{*}{Qasper} & no dict. & \R{reuse-E-noRK-qasper} & \R{merges-E-noRK-qasper} & \R{base-M-noRK-qasper} \\
 & dict. & \R{reuse-E-qasper} & \R{merges-E-qasper} & \R{base-M-qasper} \\
 & TP & \R{reuse-E-TP-qasper} & \R{merges-E-TP-qasper} & \R{base-M-TP-qasper} \\
\midrule
\multirow{3}{*}{FreshStack} & no dict. & \R{reuse-E-noRK-fs} & \R{merges-E-noRK-fs} & \R{base-M-noRK-fs} \\
 & dict. & \R{reuse-E-fs} & \R{merges-E-fs} & \R{base-M-fs} \\
 & TP & \R{reuse-E-TP-fs} & \R{merges-E-TP-fs} & \R{base-M-TP-fs} \\
\bottomrule
\end{tabular}
\caption{Key reuse and merging (exploratory). Reuse: share of chunks (\%) whose key appeared earlier in the document. Merges: chunks removed by key-based merging. TP: keys of E$-$K consolidated by a TopicGPT-style call per document, as in M$-$TP. $\Delta$: merged set minus its own unmerged enrichment (E$-$K for ``no dict.''\ and TP, E for ``dict.''), hybrid retriever, primary metric, in points.}
\label{tab:dial}
\end{table}

Table~\ref{tab:dial} compares each merged set with its own unmerged enrichment. On Qasper the loss grows with the number of merges: \R{base-M-noRK-qasper-d}, \R{base-M-qasper-d} and \R{base-M-TP-qasper-d} points. On FreshStack light merging is neutral (\R{base-M-noRK-fs}, \R{base-M-fs}), and only the heavy TP merging helps (\R{base-M-TP-fs}). Merge count and key source vary together in these rows, so a further control merges the same E$-$K chunks at M$-$TP's merge counts, at random (5 seeds) and by similarity. On FreshStack, about half of M$-$TP's gain over its base appears with any merging (random \R{dial-RM-TP-E-noRK-fs-d}, similarity \R{dial-SM-TP-E-noRK-fs-d}, consolidated keys \R{dial-M-TP-E-noRK-fs-d}), and the gain shows in recall as well as precision (\R{dial-M-TP-E-noRK-fs-rec}), so it is not only an effect of the precision metric. The consolidated keys beat random merges on both datasets (\R{dial-M-TP-RM-TP-qasper}, \R{dial-M-TP-RM-TP-fs}) but not similarity merges (\R{dial-M-TP-SM-TP-qasper}, \R{dial-M-TP-SM-TP-fs}). Across the nine merge arms, the Spearman correlation between mean chunk length and the hybrid score is \R{ax-gran-qasper-meta-rho} on Qasper and \R{ax-gran-fs-meta-rho} on FreshStack (descriptive; nine points). Two caveats remain. The random and similarity controls merge adjacent chunks only, while key-based merging can merge non-adjacent chunks; they fell \R{dial-short} merges short of the target in total because of the size cap. And prec@1024t rewards landing inside long gold sections. Consolidation failed, leaving keys unchanged, on \R{tp-failed-fs} FreshStack and \R{tp-failed-qasper} Qasper documents. E$-$K plus consolidation uses \R{cost-E-vs-TP-in-qasper}--\R{cost-E-vs-TP-in-fs}\% fewer input tokens than dictionary enrichment, but it adds one call per document and produces more output tokens.

\paragraph{Key reuse.}
Reuse is local. The median reused key was last assigned \R{ax-key-qasper-E-distmed} chunks earlier on Qasper and \R{ax-key-fs-E-distmed} on FreshStack. The 40-key cap evicted keys in \R{ax-key-qasper-E-evictdocs} Qasper and \R{ax-key-fs-E-evictdocs} FreshStack documents, and \R{ax-key-qasper-E-outside} (Qasper) and \R{ax-key-fs-E-outside} (FreshStack) of the \R{ax-key-qasper-E-reuses} and \R{ax-key-fs-E-reuses} reuses were of a key that had been evicted.

\section{Full Result Grids}
\label{app:grids}

Tables~\ref{tab:grid-qasper} and~\ref{tab:grid-freshstack} give the primary metric for every arm under every retriever; Tables~\ref{tab:grid2-qasper} and~\ref{tab:grid2-freshstack} give secondary metrics under the hybrid retriever, including the add-on arms. Per-question values for every metric, per-domain FreshStack results and the random-merge seeds are in the released results.

\begin{table*}[!htbp]
\centering
\footnotesize
\begin{tabular}{@{}lcccccccc@{}}
\toprule
\textbf{Arm} & \textbf{BM25} & \textbf{nomic} & \textbf{mxbai} & \textbf{Hybrid} & \textbf{qwen3} & \textbf{BM25+qwen3} & \textbf{Hyb.+rerank} & \textbf{jina-v3} \\
\midrule
F512c & 26.8 & 38.1 & 36.4 & 35.0 & 43.2 & 36.9 & 44.0 & -- \\
F256t & 24.1 & 37.5 & 35.5 & 35.0 & 44.6 & 38.3 & 43.6 & -- \\
FL & 23.8 & 38.4 & 35.7 & 34.3 & 42.2 & 36.1 & 42.7 & -- \\
P & 26.8 & 39.7 & 37.9 & 37.2 & 48.1 & 39.7 & 47.3 & -- \\
S & 29.7 & 47.9 & 44.4 & 42.5 & 52.0 & 44.0 & 53.0 & 46.4 \\
S+T & 29.9 & 47.8 & 47.2 & 42.7 & 53.6 & 45.2 & 54.5 & 45.9 \\
E & 38.6 & 48.6 & 47.9 & 48.1 & 55.1 & 49.7 & 54.9 & 48.7 \\
CR & 33.6 & 49.9 & 46.5 & 46.1 & 55.7 & 47.7 & 54.5 & 50.7 \\
\Estrong{} & 39.9 & 51.4 & 49.4 & 49.3 & 54.6 & 51.4 & 57.0 & -- \\
\CRstrong{} & 33.7 & 51.2 & 45.7 & 47.7 & 56.1 & 48.4 & 57.0 & -- \\
LC-S & -- & -- & -- & -- & -- & -- & -- & 35.1 \\
\bottomrule
\end{tabular}
\caption{Qasper: mean rec@512t ($\times$100) for every arm under every retriever. LC-S: late-chunked S, only under jina-v3; \CRstrong{} was built for Qasper only. Per-question values and 95\% CIs are in the released results.}
\label{tab:grid-qasper}
\end{table*}

\begin{table*}[!htbp]
\centering
\footnotesize
\begin{tabular}{@{}lccccccc@{}}
\toprule
\textbf{Arm} & \textbf{rec@256t} & \textbf{rec@512t} & \textbf{rec@1024t} & \textbf{hit@1} & \textbf{hit@3} & \textbf{hit@5} & \textbf{answer\_f1} \\
\midrule
F512c & 20.9 & 35.0 & 54.1 & 16.8 & 39.1 & 51.4 & 37.8 \\
F256t & 21.2 & 35.0 & 53.4 & 31.0 & 58.3 & 73.2 & 34.2 \\
FL & 19.2 & 34.3 & 55.1 & 26.3 & 53.4 & 69.4 & -- \\
P & 23.9 & 37.2 & 57.4 & 32.9 & 62.7 & 75.9 & -- \\
S & 27.9 & 42.5 & 62.6 & 33.2 & 60.7 & 76.7 & 37.7 \\
S+T & 28.1 & 42.7 & 62.4 & 33.9 & 62.3 & 74.4 & 37.6 \\
E & 31.7 & 48.1 & 65.0 & 39.0 & 64.2 & 77.2 & 39.0 \\
CR & 31.1 & 46.1 & 64.9 & 37.0 & 63.7 & 77.0 & 38.8 \\
\Estrong{} & 32.3 & 49.3 & 67.9 & 37.4 & 66.0 & 79.0 & 39.1 \\
\CRstrong{} & 30.6 & 47.7 & 66.4 & 34.6 & 64.8 & 78.5 & 39.0 \\
E$-$K & 32.8 & 47.5 & 65.3 & 37.5 & 64.8 & 78.2 & -- \\
M & 31.9 & 45.9 & 62.8 & 40.6 & 67.3 & 79.3 & -- \\
M$-$K & 31.6 & 46.6 & 64.9 & 37.4 & 67.3 & 79.1 & -- \\
M$-$TP & 27.3 & 41.9 & 61.4 & 41.5 & 70.9 & 83.4 & -- \\
RM & 28.8 & 44.1 & 62.5 & 37.8 & 66.7 & 79.3 & -- \\
SM & 30.3 & 46.0 & 64.1 & 38.3 & 68.7 & 81.1 & -- \\
E+KX & 29.8 & 38.2 & 53.2 & 39.0 & 53.3 & 65.9 & -- \\
E+AX & 30.4 & 40.3 & 55.3 & 39.0 & 54.0 & 65.8 & -- \\
E+SX & 32.0 & 44.3 & 59.4 & 39.0 & 58.0 & 71.0 & -- \\
E(meta+key) & 32.0 & 48.2 & 64.6 & 38.8 & 64.1 & 77.7 & -- \\
PF & 32.7 & 48.3 & 66.2 & 39.4 & 65.9 & 77.3 & -- \\
\bottomrule
\end{tabular}
\caption{Qasper: secondary metrics ($\times$100) under the hybrid retriever, including the add-on arms (M$-$K: merged E$-$K; E(meta+key): keys and entities also indexed). RM is averaged over five seeds; PF covers 100 papers.}
\label{tab:grid2-qasper}
\end{table*}

\begin{table*}[!htbp]
\centering
\footnotesize
\begin{tabular}{@{}lcccccccc@{}}
\toprule
\textbf{Arm} & \textbf{BM25} & \textbf{nomic} & \textbf{mxbai} & \textbf{Hybrid} & \textbf{qwen3} & \textbf{BM25+qwen3} & \textbf{Hyb.+rerank} & \textbf{jina-v3} \\
\midrule
F512c & 18.8 & 22.4 & 27.2 & 23.6 & 29.4 & 28.2 & 29.3 & -- \\
F256t & 23.8 & 27.8 & 30.8 & 29.5 & 34.5 & 34.6 & 35.1 & -- \\
FL & 21.1 & 22.7 & 28.5 & 25.1 & 31.8 & 30.2 & 31.8 & -- \\
P & 23.9 & 27.5 & 32.8 & 28.7 & 34.5 & 32.6 & 36.2 & -- \\
S & 22.0 & 24.2 & 30.3 & 26.7 & 31.9 & 31.1 & 31.2 & 33.7 \\
S+T & 22.7 & 26.9 & 32.2 & 28.0 & 34.6 & 32.0 & 34.1 & 33.6 \\
E & 25.1 & 29.5 & 32.2 & 30.8 & 34.8 & 33.0 & 34.9 & 32.2 \\
CR & 23.0 & 27.2 & 31.6 & 29.0 & 35.4 & 32.9 & 34.0 & 33.7 \\
\Estrong{} & 24.9 & 29.1 & 31.5 & 29.8 & 35.7 & 33.3 & 35.9 & -- \\
LC-S & -- & -- & -- & -- & -- & -- & -- & 33.2 \\
\bottomrule
\end{tabular}
\caption{FreshStack: mean prec@1024t ($\times$100) for every arm under every retriever. LC-S: late-chunked S, only under jina-v3; \CRstrong{} was built for Qasper only. Per-question values and 95\% CIs are in the released results.}
\label{tab:grid-freshstack}
\end{table*}

\begin{table*}[!htbp]
\centering
\footnotesize
\begin{tabular}{@{}lccccccc@{}}
\toprule
\textbf{Arm} & \textbf{prec@512t} & \textbf{prec@1024t} & \textbf{prec@2048t} & \textbf{rec@1024t} & \textbf{hit@1} & \textbf{hit@5} & \textbf{mrr@10} \\
\midrule
F512c & 27.4 & 23.6 & 20.1 & 6.5 & 30.2 & 61.3 & 42.9 \\
F256t & 32.4 & 29.5 & 25.3 & 8.3 & 34.0 & 60.4 & 45.2 \\
FL & 28.0 & 25.1 & 21.5 & 7.2 & 33.6 & 62.1 & 44.7 \\
P & 33.2 & 28.7 & 25.0 & 8.6 & 34.9 & 63.0 & 46.4 \\
S & 31.6 & 26.7 & 22.7 & 7.4 & 30.6 & 59.6 & 44.2 \\
S+T & 32.6 & 28.0 & 23.5 & 7.8 & 34.9 & 64.3 & 47.0 \\
E & 34.2 & 30.8 & 25.8 & 8.4 & 37.4 & 66.4 & 49.7 \\
CR & 33.2 & 29.0 & 24.3 & 8.0 & 35.7 & 63.0 & 47.5 \\
\Estrong{} & 33.1 & 29.8 & 24.7 & 8.1 & 36.6 & 65.5 & 49.1 \\
E$-$K & 33.4 & 29.4 & 24.8 & 8.2 & 36.2 & 66.0 & 49.1 \\
M & 35.4 & 31.0 & 26.4 & 8.4 & 37.9 & 66.4 & 50.0 \\
M$-$K & 32.5 & 29.9 & 24.8 & 8.4 & 34.0 & 65.1 & 47.7 \\
M$-$TP & 37.0 & 33.8 & 27.1 & 9.6 & 40.9 & 67.2 & 52.3 \\
RM & 34.8 & 30.6 & 26.0 & 8.6 & 37.8 & 66.0 & 49.7 \\
SM & 34.7 & 30.6 & 25.4 & 8.4 & 39.1 & 64.3 & 50.2 \\
E+KX & 34.4 & 32.0 & 27.1 & 8.7 & 37.4 & 52.3 & 43.6 \\
E+AX & 36.2 & 33.5 & 28.6 & 9.3 & 37.4 & 50.6 & 42.5 \\
E+SX & 35.1 & 31.5 & 27.1 & 8.6 & 37.4 & 60.4 & 47.7 \\
E(meta+key) & 33.7 & 30.3 & 26.1 & 8.3 & 36.6 & 67.2 & 49.3 \\
\bottomrule
\end{tabular}
\caption{FreshStack: secondary metrics ($\times$100) under the hybrid retriever, including the add-on arms (M$-$K: merged E$-$K; E(meta+key): keys and entities also indexed). RM is averaged over five seeds.}
\label{tab:grid2-freshstack}
\end{table*}

\section{Cost Model}
\label{app:cost}

Header-led chunking and key-based merging make no model calls. Parsing and grouping take time linear in the document length $|D|$; merging adds an $O(n \log n)$ sort and linear-time packing and concatenation. The cost of indexing is the cost of the enrichment calls.

Measure prompt length in tokens. Let $P$ be the fixed instruction text of the enrichment prompt, $t_i$ the length of chunk $c_i$, $h_i$ its section path, $s_{i-1}$ the previous summary, $\kappa$ the mean length of one rendered key line, and $o_i$ the length of the JSON response. One document costs $n$ calls (in the normal case) with
\begin{align}
  T_{\text{in}} &= \sum_{i=1}^{n} \bigl( P + t_i + h_i + s_{i-1} + \kappa\,|K_{i-1}| \bigr) \nonumber\\
  &\leq n\,(P + \bar{h} + \bar{s} + \kappa \Kmax) + \textstyle\sum_i t_i, \label{eq:single}\\
  T_{\text{out}} &= \textstyle\sum_{i=1}^{n} o_i . \nonumber
\end{align}
The rolling dictionary adds at most $\kappa\Kmax$ tokens per call. Because $P$ (the full instruction text) is of the same order as a chunk of $\taumax = 1{,}500$ characters, the fixed overhead is a large share of each call.

A pipeline that runs $m$ separate extractors, each with its own instruction text $P_j$ and the chunk, costs $nm$ calls and
\begin{equation}
  T_{\text{in}}^{(m)} = n \sum_{j=1}^{m} P_j \;+\; m \sum_i t_i ,
  \label{eq:multi}
\end{equation}
with roughly the same total output, since the same fields are produced. The single call sends each chunk once instead of $m$ times and pays one instruction overhead instead of $m$. Contextual retrieval also makes one call per chunk but sends the whole document with each chunk, so its input is about $n\,(P_{\text{CR}} + |D|) + \sum_i t_i$, which grows quadratically with document length unless the document prefix is cached.

Calls within a document are sequential, because call $i$ needs $K_{i-1}$ and the summary of $c_{i-1}$. Documents are independent, so different documents can be enriched in parallel. In the harness, a chunk whose response cannot be parsed is retried up to three times (Appendix~\ref{app:repro}). Measured costs are in Table~\ref{tab:pilotcost}.

\paragraph{Confirmatory costs.} Per call, CR sent \R{crratio-local-qasper} (Qasper) and \R{crratio-local-fs} (FreshStack) times the input tokens of enrichment with the 7B indexing LLM, and \R{crratio-strong-qasper} times with the 120B indexing LLM on Qasper. In total, local enrichment sent \R{tok-local-enr-qasper-in} (Qasper) and \R{tok-local-enr-fs-in} (FreshStack) input tokens, and local CR \R{tok-local-cr-qasper-in} and \R{tok-local-cr-fs-in}. Measured API spend for the 120B indexing LLM was US\$\R{usd-enr-qasper} for enrichment and US\$\R{usd-cr-qasper} for CR on Qasper, and US\$\R{usd-enr-fs} for enrichment on FreshStack. Per call this is \R{usdpercall-enr-qasper} micro-dollars for enrichment and \R{usdpercall-cr-qasper} for CR on Qasper, and \R{usdpercall-enr-fs} for enrichment on FreshStack. The per-call price gap (about twice) is far smaller than the token ratio because input tokens are cheap at this provider and enrichment returns longer outputs (\R{tok-strong-enr-qasper-outpercall} against \R{tok-strong-cr-qasper-outpercall} tokens per call; Table~\ref{tab:pilotcost}). Total API spend for the study was US\$\R{spend}.

Table~\ref{tab:pilotcost} gives the costs measured in the pilot from Ollama's per-call counters (\texttt{prompt\_eval\_count}, \texttt{eval\_count}). Showing the model the rolling dictionary added about 110 input tokens per call (1,002 for E against 892 without the dictionary). This matches the term $\kappa\,|K_{i-1}|$ of Eq.~\ref{eq:single}. Wall-clock times were measured under four-way concurrency on one GPU with prefix caching, so they do not isolate the cost of a configuration, and we do not compare them.

\begin{table}[!htbp]
\centering
\footnotesize
\setlength{\tabcolsep}{2pt}
\begin{tabular}{@{}lrrrr@{}}
\toprule
 & \multicolumn{2}{c}{\textbf{Qasper}} & \multicolumn{2}{c}{\textbf{FreshStack}} \\
\textbf{Arm} & \textbf{In} & \textbf{Out} & \textbf{In} & \textbf{Out} \\
\midrule
\multicolumn{5}{@{}l}{\emph{Pilot}} \\
Zero-call arms & 0 & 0 & 0 & 0 \\
CR (7B) & 5{,}520 & 58 & 9{,}825 & 48 \\
E (7B) & 1{,}002 & 188 & 1{,}065 & 168 \\
E without dictionary & 892 & 188 & -- & -- \\
\midrule
\multicolumn{5}{@{}l}{\emph{Confirmatory}} \\
E (7B) & \R{tok-local-enr-qasper-percall} & \R{tok-local-enr-qasper-outpercall} & \R{tok-local-enr-fs-percall} & \R{tok-local-enr-fs-outpercall} \\
CR (7B) & \R{tok-local-cr-qasper-percall} & \R{tok-local-cr-qasper-outpercall} & \R{tok-local-cr-fs-percall} & \R{tok-local-cr-fs-outpercall} \\
\Estrong{} & \R{tok-strong-enr-qasper-percall} & \R{tok-strong-enr-qasper-outpercall} & \R{tok-strong-enr-fs-percall} & \R{tok-strong-enr-fs-outpercall} \\
\CRstrong{} & \R{tok-strong-cr-qasper-percall} & \R{tok-strong-cr-qasper-outpercall} & -- & -- \\
\bottomrule
\end{tabular}
\caption{Mean input and output tokens per LLM call. Pilot rows come from the pilot results files (one call per header-led chunk: 715 on Qasper, 742 on FreshStack). Confirmatory rows come from the server's token counters. Output tokens of the 120B indexing LLM are as billed by the provider.}
\label{tab:pilotcost}
\end{table}

\section{Pilot (Exploratory)}
\label{sec:pilot}

The pilot used 30 Qasper papers (79 questions) and 73 FreshStack-Laravel questions. Its plan asked RQ1 (header-led chunks against fixed windows, the later H3), RQ2 (E against S+T), RQ3 (E against CR) and RQ4 (rolling keys: 4a key reuse; 4b retrieval of merged chunks with and without the dictionary, and merged against unmerged); its arm E+M is E with key-based merging (the later M), and E$-$K+M is E$-$K with key-based merging. We report it because it shaped the confirmatory plan. It is exploratory throughout: the pilot plan was not pushed before results (\S\ref{sec:plan}), and every re-analysis below was run after the pilot results had been viewed. Table~\ref{tab:pilot} restates the planned pilot cells, recomputed with 10,000 resamples. The pilot used a wrong Laravel commit. Correcting it moves the FreshStack RQ1 cells (S $-$ F512c, S $-$ F256t) by at most 0.3 points and the enrichment cells by up to 1.9 points; Table~\ref{tab:pilot} shows the corrected FreshStack values (Appendix~\ref{app:erratum}).

\paragraph{Paid enrichment.}
Under hybrid retrieval, every E $-$ S+T and E $-$ CR cell is \emph{inconclusive}: none is equivalent and none survives Holm adjustment. The only equivalent enrichment cell is FreshStack BM25 E $-$ S+T after the erratum correction. For example, the 90\% CI for Qasper hybrid E $-$ S+T is [$-$7.6, +3.2], which does not fit inside $\pm$5. The pilot therefore cannot say whether enrichment helps. This is why the confirmatory samples are 3--12 times larger. Under BM25 alone, E $-$ S+T on Qasper was positive in all 12 BM25 settings we tried (per-paper or corpus IDF, three $k_1$, two $b$), from +4.1 to +8.8 points; its CI excluded 0 in 2 of them. A field ablation attributes most of the gain to the generated questions: S+questions $-$ S was +6.1 [$-$0.6, +13.1], against +8.0 for all fields. This is consistent with a doc2query-style lexical effect, as \citet{weller2024expansions} predict. Under hybrid retrieval, indexing only the title and summary \emph{lowered} recall ($-$7.0 [$-$12.6, $-$1.5]). The secondary embedders gave conflicting signs. On FreshStack, enrichment helped mxbai (+4.7 [+0.3, +8.9] over S+T, corrected commit). That gain came from placing the metadata before the chunk text. Truncation does not explain it: placing the metadata after the text lowered the score by 3.5 [$-$6.3, $-$0.6], and only 1--2.4\% of chunks exceed mxbai's 512-token limit. Under qwen3-embedding, E $-$ S+T was $-$3.3 [$-$8.1, +1.3] on FreshStack and +9.2 [+0.1, +17.5] on Qasper.

\paragraph{Structure and alignment.}
One objection is that S wins on Qasper only because S never splits an evidence paragraph. F512c splits 83.9\% of evidence paragraphs and F256t 37.1\%. A greedy oracle can reach rec@512t of 0.94--0.99 with every chunk set, while the best observed score is about 0.50. Retrieval is therefore limited by ranking, not by what a chunk set can reach. The header-blind paragraph packer P keeps every paragraph whole, yet scores the same as F256t (hybrid +0.3 [$-$11.1, +11.4]). S beats P by +12.4 [+5.1, +20.3] under hybrid retrieval. Against length-matched windows averaged over four window offsets, S gains +15.4 [+8.8, +22.8] under hybrid retrieval and +8.0 [+1.9, +14.2] under BM25. In the pilot, the gain follows header-bounded sections, not whole paragraphs. One caution applies: the BM25 score of a single fixed-window setting ranged from 0.20 to 0.32 across window sizes and offsets.

\paragraph{Keys and merging.}
With the rolling dictionary, key reuse rose from 5.5\% to 14.7\% of keyed Qasper chunks. The one ``significant'' planned pilot cell outside RQ1 (E+M $-$ E$-$K+M under BM25, $-$6.0 [$-$12.8, $-$0.2]) has a permutation $p$ of 0.092 and does not survive Holm adjustment. We do not treat it as a finding. Key-based merges beat random merges under hybrid retrieval (+4.8 [+0.5, +9.7]) but did not beat no merging.

\begin{table*}[!htbp]
\centering
\footnotesize
\setlength{\tabcolsep}{5pt}
\begin{tabular}{@{}lcccc@{}}
\toprule
 & \multicolumn{2}{c}{\textbf{Qasper} (rec@512t)} & \multicolumn{2}{c}{\textbf{FreshStack} (prec@1024t)} \\
\textbf{A $-$ B} & \textbf{Hybrid} & \textbf{BM25} & \textbf{Hybrid} & \textbf{BM25} \\
\midrule
S $-$ F512c & \textbf{+23.0} {\scriptsize[+12.9,+33.5]} & +9.9 {\scriptsize[+0.6,+18.3]} & +4.9 {\scriptsize[+1.2,+8.8]} & +3.0 {\scriptsize[+0.0,+5.9]} \\
S $-$ F256t & \textbf{+12.7} {\scriptsize[+5.5,+20.2]} & +4.6 {\scriptsize[$-$3.6,+12.7]} & $-$2.3 {\scriptsize[$-$6.2,+1.4]} & $-$2.2 {\scriptsize[$-$6.0,+1.5]} \\
E $-$ S+T & $-$2.3 {\scriptsize[$-$8.6,+4.3]} & +6.8 {\scriptsize[$-$0.8,+14.8]} & +0.9 {\scriptsize[$-$2.9,+4.3]} & $-$0.1 {\scriptsize[$-$3.0,+2.9]}$^{\equiv}$ \\
E $-$ CR & $-$4.1 {\scriptsize[$-$10.4,+2.3]} & +2.4 {\scriptsize[$-$5.6,+10.4]} & $-$0.2 {\scriptsize[$-$3.9,+3.7]} & $-$0.5 {\scriptsize[$-$3.7,+2.8]} \\
E+M $-$ E & +2.4 {\scriptsize[$-$0.2,+5.9]} & $-$3.1 {\scriptsize[$-$9.3,+3.1]} & +0.5 {\scriptsize[$-$1.1,+2.1]}$^{\equiv}$ & +0.5 {\scriptsize[$-$1.8,+2.7]}$^{\equiv}$ \\
\bottomrule
\end{tabular}
\caption{Pilot (exploratory). Planned paired differences in points, with 95\% percentile cluster-bootstrap CIs. Bold: survives Holm adjustment over all 22 planned pilot cells. $^{\equiv}$: equivalent within $\pm$3 points (TOST). FreshStack cells use the corrected Laravel commit and, for E, CR and E+M, the rebuilt indexes (Appendix~\ref{app:erratum}). Apart from FreshStack BM25 E $-$ S+T after the correction, no E $-$ S+T or E $-$ CR cell is equivalent within $\pm$5 (Qasper) or $\pm$3 (FreshStack).}
\label{tab:pilot}
\end{table*}

\subsection{Corrected Laravel Commit}
\label{app:erratum}

The pilot located FreshStack gold spans in \texttt{laravel/docs} at commit \texttt{1e8496c2} (2024-10-30). FreshStack's corpus instead matches \texttt{724c31cc} (2024-10-24). Over the full docs root, 26 corpus chunks in 7 files differ between the two commits. Among the 61 unique pilot gold spans, 6 were not byte-exact. Five, in \texttt{eloquent-relationships.md}, came from the wrong commit. One, in \texttt{releases.md}, came from a second bug: FreshStack offsets count characters, the pilot sliced bytes, and two non-ASCII characters precede that span. The 6 spans affect 15 of the 73 pilot questions. We re-sliced all spans by character offset at the correct commit (61 of 61 now exact) and rebuilt the one changed indexed file for every arm, including the LLM arms E, E+M and CR. Table~\ref{tab:erratum} gives the planned RQ1 cells; Table~\ref{tab:pilot} gives all corrected FreshStack pilot cells. The RQ1 cells move by at most 0.3 points and the enrichment cells by up to 1.9 points. No conclusion changes.

\begin{table}[!htbp]
\centering
\footnotesize
\scriptsize
\setlength{\tabcolsep}{2.5pt}
\begin{tabular}{@{}llcc@{}}
\toprule
\textbf{Retr.} & \textbf{A $-$ B} & \textbf{Wrong commit} & \textbf{Corrected} \\
\midrule
Hybrid & S $-$ F512c & +5.1 [+1.4, +9.0] & +4.9 [+1.2, +8.8] \\
Hybrid & S $-$ F256t & $-$2.6 [$-$6.4, +1.2] & $-$2.3 [$-$6.2, +1.4] \\
BM25 & S $-$ F512c & +3.0 [$-$0.0, +6.0] & +3.0 [+0.0, +5.9] \\
BM25 & S $-$ F256t & $-$2.2 [$-$6.0, +1.5] & $-$2.2 [$-$6.0, +1.5] \\
\bottomrule
\end{tabular}
\caption{FreshStack pilot RQ1 cells (prec@1024t, points, 95\% bootstrap CI over the 73 questions) before and after correcting the Laravel commit. The left column comes from the original bootstrap, so its CIs can differ by 0.1 from a recomputation.}
\label{tab:erratum}
\end{table}

\section{Plan History, Amendments and Deviations}
\label{app:plan}

\paragraph{Pilot plan.}
The pilot plan (\commit{084b45d}, 2026-10-02 15:51 IST) fixed the primary metrics, the primary retrievers (hybrid and BM25) and six paired differences. It was amended at 15:59 (\commit{9e37ea4}, \commit{41eb775}) after a harness audit. The audit's fixes changed no metric, budget, retriever or comparison. They added paired CIs for every planned comparison. They took n-grams within each retrieved chunk, since concatenation had penalized small chunks. They dropped section-name evidence and required short evidence to match contiguously. They computed key reuse over keyed chunks on paired documents. They also changed the ablation prompt, which had contradicted the chunk position, and rebuilt that arm with seed 0. Key-reuse statistics from the flawed ablation prompt had been viewed on about 18 partially built papers before the fix. The commit that added the plan also contained descriptive chunk statistics for 21 papers and key and cost statistics for 18--19 papers. The pilot plan, its commit message and the pilot harness comments call it ``pre-registered''. Because it was committed only locally and pushed after results, we do not use that term.

\paragraph{Comparison counts.}
The pilot plan's text counted ``4 primary comparisons'' and expected about 0.8 false positives. Counted as paired differences, it specified six (RQ1 has two, RQ4b has two). With RQ1--RQ3 on both datasets, RQ4b on Qasper only, and two primary retrievers, that makes 20 primary-retriever intervals, with about 1.0 expected to exclude 0 under the global null. The Phase~0 re-analysis (six zero-cost re-analyses of the pilot, run before the confirmatory plan) applies Holm adjustment over 22 cells, which adds the two descriptive FreshStack E+M $-$ E cells. The confirmatory plan has a single primary retriever and explicit families: a primary family of four tests, an H3 family of four, and a key-secondary family.

\paragraph{What had been viewed before the confirmatory plan.}
Before writing the confirmatory plan, we had viewed every pilot result, six zero-cost re-analyses of the pilot (Phase~0), descriptive statistics of the new samples, an AI-assisted mock review of an earlier draft, and a literature check. No retrieval, answer or judge metric had been computed on the new samples.

\paragraph{Confirmatory choices informed by the pilot.}
\begin{enumerate}[leftmargin=*,label=(\roman*)]
  \item BM25 was demoted from co-primary to secondary, to reduce multiplicity.
  \item The margins were set from the pilot's paired SDs. The FreshStack margin is $\pm$4 rather than $\pm$3 because excluding the pilot questions leaves 235 questions, for which a $\pm$3 margin gives about 0.65 power.
  \item The P, FL and random-merge controls came from the Phase~0 re-analyses.
  \item The Qasper section-path filter and the FreshStack commit identification came from auditing the pilot data.
  \item The gold-plus-distractor cap for Angular and YOLO was set from descriptive chunk counts to fit compute. Laravel was kept in full.
\end{enumerate}

\paragraph{Confirmatory plan.}
The plan (\commit{dd1eedb}) was committed at 13:45:06 IST on 2026-10-06 and pushed to GitHub at 13:45:16 IST; GitHub's server-side push record provides the external timestamp. The frozen configuration was written at 13:46:30 IST and the first build started at 13:46.

\paragraph{Amendment 1} (\commit{67c50e9}, committed 14:04:42 IST and pushed 14:04:45 IST on 2026-10-06). It was made while builds were running and before any outcome metric was computed. It changed nothing in plan sections 1--5 and added only secondary or exploratory analyses with unadjusted CIs:
\begin{enumerate}[leftmargin=*,label=(\Alph*)]
  \item A Qwen3-Reranker-0.6B reranking of the hybrid top-30 (fallback: bge-reranker-v2-m3); qwen3-embedding as a dense retriever and in a BM25 hybrid; and late chunking with jina-embeddings-v3, pooled over header-led chunk spans and compared with plain jina-v3 on S, S+T, E and CR.
  \item FreshStack strict nugget recall, judged by \texttt{llama-3.3-70b-instruct} (DeepInfra, temperature 0). The plan's ``no LLM judge'' rule was replaced by this secondary outcome, with acceptance controls and a $\pm$5 margin.
  \item An exploratory scaling arm: E with Qwen3-235B-A22B-Instruct-2507 on 30 seeded Qasper papers, of which 23 were completed (see Deviations).
  \item A re-scoring of the pilot on the correct Laravel commit (Appendix~\ref{app:erratum}).
\end{enumerate}

\paragraph{Amendments 2--4 (2026-10-08).}
All three were written after the H1--H3 results had been computed and viewed, so every test in them is pre-specified secondary. Each was pushed before the arms it governs were built or scored. The confirmatory plan and earlier amendments were left unchanged.
\begin{enumerate}[leftmargin=*,label=(\arabic*)]
  \setcounter{enumi}{1}
  \item Amendment~2 (\commit{0502ffe}, pushed 13:36:29 IST; a wording correction to the ablation placeholder followed in \commit{a46c5e4}, before any build). It added the merge arms M, RM and SM, the dictionary ablation E$-$K (also with key-based merging applied), and the H4 family (H4a--d, eight tests), with the decision rule that \emph{diff.}\ also requires a Holm-adjusted sign-flip $p<0.05$. The merge arms were built after 13:36 and the dictionary ablation from 16:04.
  \item Amendment~3 (\commit{fa2c215}, pushed 13:45:36 IST). It added related-key expansion E+KX with the count-matched adjacency control E+AX (H5a--b, four tests), the per-field arm PF on 100 Qasper papers (H6), and exploratory indexing of keys and entities. The related-key arms were scored after 13:45.
  \item Amendment~4 (\commit{34ad2b6}, pushed 13:49:07 IST). Prompted by a literature check made before any Amendment~4 arm existed, it added the TopicGPT-style consolidation arm M$-$TP (H7, two tests) and exploratory section expansion E+SX. It was written before E$-$K was built and before any H5 outcome was computed.
\end{enumerate}

\paragraph{Planned analyses not run.}
Section~6 of the plan document listed three secondary analyses that were not run: robustness runs (repeat generations and a paraphrased prompt), pooled-corpus Qasper (retrieval across all sampled papers rather than within each paper), and field ablations of the enrichment output. No result in this paper therefore measures variation between generation runs or prompts, or cross-paper Qasper retrieval, and the contribution of individual fields is known only from the exploratory pilot ablation (Appendix~\ref{sec:pilot}). The per-domain FreshStack results, also listed there, are reported for H1 (\S\ref{sec:confirm}).

\paragraph{Deviations.}
The study log (\texttt{DEVIATIONS.md}) records the following; none changed an arm, prompt, model or decoding setting of a confirmatory test.
\begin{enumerate}[leftmargin=*,label=(\roman*)]
  \item \emph{Scaling arm truncated.} The exploratory Qwen3-235B arm ran on a provider's free quota. It stopped at the token cap with 23 of the 30 sampled papers complete and no failures; the stopping point depends only on processing order. The comparison uses those 23 papers.
  \item \emph{Interim view during a smoke test.} A smoke test of the secondary-retriever code printed rec@512t for S and S+T on one Qasper paper (4 questions) under the hybrid, qwen3 and BM25+qwen3 retrievers. No enrichment or CR arm and no paired comparison was involved; the output was deleted, and later runs suppress statistics until all builds finish.
  \item \emph{FreshStack local build split for throughput.} Enrichment ran at context length 8,192 with eight parallel slots and CR at 32,768 with four. The worst-case enrichment prompt fits in 8,192 tokens, so outputs are unaffected. No FreshStack local output existed before the change.
  \item \emph{qwen3-embedding reruns.} The first run stopped when the local model server was down and wrote no output; it was rerun with identical settings into a separate folder. The settings below were fixed on 12--13 October, after the H1--H3 results had been viewed (by 8 October) but before any qwen3 statistic was computed. The embedding runner then crashed on seven long chunk texts in one Qasper paper: it splits each input into micro-batches of 2,048 tokens and crashes, rather than truncating, on any input that spans two. The arm therefore loads the model with an 8,192-token context and a single 8,192-token micro-batch, retries rejected batches one text at a time, and cuts texts and queries longer than 3,000 Qwen3 tokens to 3,000; embeddings of shorter texts were checked to be bit-identical. A further crash on three long FreshStack queries required a rerun of that dataset only.
  \item \emph{Late chunking out of memory.} The late-chunking run stopped on one Qasper paper when a batch of eight long texts exceeded GPU memory. A batch that runs out of memory is now re-encoded one text at a time; on 16 checked texts, batch-of-one and batch-of-eight embeddings agree to cosine $\ge$0.99996. Cached embeddings were reused.
  \item \emph{Pauses.} Runs were paused three times by the operator and resumed where they stopped; outputs are unaffected.
  \item \emph{Pilot erratum re-scored.} After the rebuild of the changed Laravel file, the LLM arms of the pilot were re-scored (Appendix~\ref{app:erratum}).
  \item \emph{Corrections from a separate re-computation.} A re-computation from the per-question outputs matched every confirmatory and add-on difference and CI available on 12 October. It found reporting errors, which were corrected without new data: verdict labels now apply each family's rule (\S\ref{sec:stats}), which changed Qasper M $-$ RM and Qasper E+KX $-$ E+AX from \emph{diff.}\ to \emph{n.d.}\ and FreshStack \Estrong{} $-$ S+T from \emph{equiv.}\ to \emph{incon.}; the key-secondary family was restricted to the five tests the plan specifies; four FreshStack judgements that stayed unparseable were counted as unsupported, as the plan requires, rather than dropped; and wall-clock times per field are no longer reported.
\end{enumerate}

\section{Dataset Construction}
\label{app:data}

Both preparation scripts make no LLM or API calls. Tokens are counted with \texttt{cl100k\_base}.

\paragraph{Qasper.}
The pool is the 697 dev and test papers minus the 30 pilot papers, which leaves 667. A paper is eligible if at least one question has text evidence that does not start with ``FLOAT SELECTED'' and is not a section path (strings containing ``\texttt{:::}''). The section-path filter is new; the pilot dropped such strings only from the evidence sets. Evidence from annotators who marked a question unanswerable is ignored. Of the 667 papers, 654 are eligible. The pool is sorted by paper id, and \texttt{random.Random(2027).sample} draws 300 of them (105 dev, 195 test). This gives 951 questions. By number of annotators, 80 questions have 1, 601 have 2, 233 have 3, 33 have 4 and 4 have 5. The 2,245 evidence paragraphs come to 1, 2, 3 and 4+ per question for 432, 269, 100 and 150 questions. Of these paragraphs, 2,231 are found verbatim in the Markdown; the other 14 differ only in whitespace or encoding. The Markdown totals 1.45M tokens and header-led chunking yields 7,222 chunks. Each question stores both the union of evidence over annotators (primary) and per-annotator evidence (for a max-over-annotators secondary analysis).

\paragraph{FreshStack.}
FreshStack's corpus metadata gives only branch URLs. For each candidate repository commit in the release window, we counted how many corpus chunks under the docs root match that commit's file text exactly. Table~\ref{tab:fscommits} lists the commit with the most matches. At each chosen commit, the set of \texttt{.md} files under the root equals the corpus file set. Every file is fetched at that commit and checked against the git blob SHA-1. Gold is the union of the relevant corpus ids over all nuggets of a question (\texttt{queries-oct-2024}, test split). A span is kept if its corpus text equals the Markdown substring at its offsets. All kept spans were verified byte-exact. Non-Markdown spans and spans outside the docs root are dropped and counted for each question. Questions with no kept span are dropped. The selection, including the distractor files (\texttt{random.Random(2027)}, ratio 0.5, Angular and YOLO only), was committed with the plan.

\begin{table*}[!htbp]
\centering
\footnotesize
\setlength{\tabcolsep}{4pt}
\begin{tabular}{@{}llrrrrrr@{}}
\toprule
\textbf{Domain} & \textbf{Commit (date)} & \textbf{Chunks matched} & \textbf{.md files} & \textbf{Doc tokens} & \textbf{S chunks} & \textbf{Questions kept} & \textbf{Gold files} \\
\midrule
Laravel & \texttt{724c31cc} (2024-10-24) & 455/455 & 98 & 644k & 3{,}860 & 89 & 71 \\
Angular & \texttt{7d9b38e9} (2024-10-28) & 383/384 & 283 & 387k & 2{,}307 & 99 & 118 \\
YOLO & \texttt{8d203cf4} (2024-10-26) & 565/565 & 335 & 622k & 3{,}893 & 47 & 68 \\
\bottomrule
\end{tabular}
\caption{FreshStack corpora at the identified commits (counts before the distractor cap). Questions: Laravel 184 total, 22 without Markdown gold, 73 pilot questions excluded; Angular 129 total, 30 without Markdown gold; YOLO 57 total, 10 without Markdown gold. The single unmatched Angular chunk is in \texttt{image-optimization.md}, where FreshStack removed an inline base64 image; no kept gold span is affected. Of the 235 kept questions, 163 have all their Markdown gold under the docs root.}
\label{tab:fscommits}
\end{table*}

\section{Reproducibility and Responsible NLP Checklist}
\label{app:repro}

\paragraph{Frozen configuration.}
\texttt{results/v2/frozen\_config.json} was written at 13:46:30 IST on 2026-10-06. Builds refuse to run if the configuration differs from it. It records the following. Local model \texttt{qwen2.5:7b}, Ollama digest \texttt{845dbda0\ldots}, temperature 0, seed 0, context length 16,384 (Qasper) or 32,768 (FreshStack; 8,192 for FreshStack enrichment after the split build, see Deviations), output limit 1,000 tokens for enrichment and 120 for CR. 120B indexing LLM \texttt{openai/gpt-oss-120b}, pinned to provider \texttt{deepinfra/bf16} with fallbacks disabled and data collection denied, low reasoning effort, temperature 0, seed 0, output limit 2,000 tokens for enrichment and 600 for CR. SHA-256 hashes of the enrichment, CR and answer prompts. Chunking parameters. The harness and \system{} commits (\commit{dd1eedb}, \commit{b5b6f3c}, both clean). \system{} version \sysversion{}. LLM parse failures are retried with seeds 1--3, after which the chunk keeps empty fields and is counted.

\paragraph{Released artifacts.}
The release includes the harness, both analysis plans\ifreview\else{} with their full git history\fi{}, the four amendments, the selection files, the frozen configuration, the spend ledger, and the per-question results for every arm and retriever, \repo{}. The generated enrichment and CR texts (about 535\,MB; 95\,MB compressed) \cacherelease{}.

\paragraph{ACL Responsible NLP checklist.}
\begin{description}[leftmargin=0pt,style=unboxed,itemsep=1pt]
  \item[A1 Limitations:] yes, in the unnumbered Limitations section.
  \item[A2 Risks:] yes, in the Ethics Statement. The study evaluates retrieval and poses low direct risk.
  \item[B Artifacts used:] Qasper and FreshStack (\S\ref{sec:data}), Laravel, Angular and Ultralytics documentation, and the models listed above. Licenses: Qasper CC BY 4.0; FreshStack queries and corpora CC BY-SA 4.0 (Stack Overflow content is CC BY-SA 4.0); corpus repositories MIT, BSD-3-Clause, Apache-2.0, CC BY 4.0 (Angular documentation), GPL-3.0 and AGPL-3.0 (Ultralytics); Qwen2.5, Qwen3, gpt-oss, nomic-embed-text and mxbai-embed-large Apache-2.0; Llama~3.3 under the Llama~3.3 Community License; jina-embeddings-v3 CC BY-NC 4.0 (non-commercial). Our use is non-commercial research evaluation, consistent with all of these; released files that contain corpus text keep the source licences, including ShareAlike.
  \item[B Artifacts created:] the harness and data selection files, released with \system{} under the MIT license.
  \item[B Data statistics:] Appendix~\ref{app:data}.
  \item[B Personal information:] Stack Overflow questions are public posts. We add no personal data and use no identifiers beyond question ids.
  \item[C Computation:] local inference on one Apple M5 laptop (24\,GB); API generation through OpenRouter (120B indexing LLM), DeepInfra (judge) and a free DashScope quota (exploratory scaling arm); spend capped at US\$5.50, actual spend US\$\R{spend}. All local computation ran between 6 and 16 October 2026 on that laptop, including operator pauses; GPU-hours were not metered.
  \item[C Hyperparameters:] all fixed in advance (Appendix~\ref{app:repro}); prompts in Appendix~\ref{app:prompt} and the release; no tuning on test data.
  \item[C Statistics:] BCa cluster bootstrap with 10,000 resamples, TOST, Holm adjustment, permutation check (\S\ref{sec:stats}).
  \item[C Packages:] Python 3.12.13, rank\_bm25 0.2.2, Ollama 0.32.5, sentence-transformers 5.7.0, transformers 4.57.6, PyTorch 2.14.1, tiktoken 0.14.0, NumPy 2.5.3, SciPy 1.18.1.
  \item[D Human annotation:] none; the optional spot-check of judge labels was not completed.
  \item[E AI assistants:] used for coding and for editing the text; the authors checked all code, numbers and text.
\end{description}

\section{How Indexing Methods Spend LLM Calls}
\label{app:comparison}
Table~\ref{tab:comparison} compares how common indexing methods place boundaries, what text they add, what context the model sees and how many LLM calls they make per chunk.

\begin{table*}[!htbp]
\centering
\footnotesize
\setlength{\tabcolsep}{3pt}
\renewcommand{\arraystretch}{1.15}
\begin{tabularx}{\linewidth}{@{}>{\raggedright\arraybackslash}p{2.65cm} Y Y Y Y >{\raggedright\arraybackslash}p{1.6cm}@{}}
\toprule
\textbf{Method} & \textbf{Boundaries} & \textbf{Added text} & \textbf{Context seen} & \textbf{Merging} & \textbf{LLM calls} \\
\midrule
Fixed-size & Char./token window & None & -- & None & 0 \\
Meta-Chunking\newline\citep{zhao2024metachunking} & LM perplexity / margin sampling & Chunk rewriting (later version) & Neighbouring sentences & Combine to target length & LM scoring per sentence \\
LumberChunker\newline\citep{duarte2024lumberchunker} & LLM-detected topic shift & None & Group of passages & None & $\approx$1 per boundary \\
Element-type\newline\citep{jimenoyepes2024financial} & Document elements & None & -- & None & 0 \\
Title chain\newline\citep{yang2026titlechain} & Markdown headers & Section-title prefix & Section path & None & 0 \\
Late chunking\newline\citep{gunther2024late} & Any & None (contextual embeddings) & Whole document (embedder) & None & 0 \\
Doc2query\newline\citep{nogueira2019doc2query} & Passage & Predicted questions & Passage only & None & 1 (seq2seq model) \\
Per-field extractors & Any & Title, summary, keywords, questions & Varies & None & $m$ (one per field) \\
Contextual retrieval\newline\citep{anthropic2024contextual} & Any & Situating context & Whole document & None & 1 \\
RAPTOR\newline\citep{sarthi2024raptor} & Fixed ($\approx$100 tokens) & Cluster summaries (new nodes) & Cluster members & Adds summary tree & $<$1 \\
GraphRAG\newline\citep{edge2024graphrag} & Fixed token units & Entities, relations, claims; community summaries & Unit; community & Graph + communities & $\geq$1 \\
CDTA\newline\citep{stankovic2025cdta} & Topic segments & Synthesized topic chunks & Segments across documents & Cross-document synthesis & -- \\
\midrule
Per-chunk enrichment (E; this study) & Header-led & Title, summary, keywords, questions & Chunk, section path, previous summary & None & 1 \\
\quad + rolling key dictionary & Header-led & Same as E & + up to 40 earlier keys of the document & None & 1 \\
\quad + key-based merging (M) & Header-led & Same as E & Same as E & Same-key concatenation within a document & 1 \\
\quad + related-key expansion (E+KX) & Header-led & Same as E; linked chunks added at retrieval & Same as E & None & 1 \\
\bottomrule
\end{tabularx}
\caption{How indexing methods place boundaries, what text they add, what context the model sees, how they merge or regroup chunks, and how many LLM calls they make per chunk. The last block lists the technique tested here and its optional add-ons (\S\ref{sec:method}). ``--'' means not applicable or not comparable per chunk.}
\label{tab:comparison}
\end{table*}

\section{Implementation of the Indexing Arms}
\label{app:method}

All header-led arms are built with \system{} (release \sysversion{}), called from the evaluation harness, whose code is \repo{}. This appendix gives the rules needed to reproduce the arms: header-led chunking, the enrichment call and the three optional add-ons tested here (rolling key dictionary, key-based merging, related keys). Figure~\ref{fig:pipeline} shows how they fit together. The harness calls the enricher through its own client, with temperature 0, seed 0, JSON output and up to three parse retries with seeds 1--3 (Appendix~\ref{app:repro}). Results therefore describe the harness's LLM settings (model, temperature 0, seed 0, output limits), not the package's LLM defaults (such as temperature 0.1), which were not evaluated; chunking uses the package defaults, and key-based merging is applied only in the merge arms; the package's configuration options, output schema and chunk identifiers are documented in the repository.

\begin{figure*}[!htbp]
\centering
\resizebox{0.8\textwidth}{!}{%
\begin{tikzpicture}[
  font=\small,
  box/.style={draw, rounded corners=2pt, align=center, minimum height=1.25cm, text width=2.55cm, inner sep=3pt},
  io/.style={draw, align=center, minimum height=1.25cm, text width=1.65cm, inner sep=3pt},
  arr/.style={-{Stealth[length=2mm]}, thick},
  lab/.style={font=\scriptsize, align=center}
]
\node[io] (doc) {Markdown\\document $D$};
\node[box, right=0.55cm of doc] (s1) {\textbf{Header-led chunking}\\parse blocks;\\group by section};
\node[box, right=1.1cm of s1, fill=gray!8] (s2) {\textbf{Enrichment}\\for $i=1..n$:\\one LLM call for $c_i$};
\node[box, right=1.25cm of s2, dashed] (s3) {\textbf{Key-based merging} (optional)\\group by key;\\next-fit packing};
\node[io, right=0.55cm of s3] (out) {indexed\\chunks};
\node[box, below=0.8cm of s2, xshift=-1.55cm, text width=2.7cm, dashed] (K) {Rolling key dictionary $K$ (optional; at most $\Kmax$ keys, LRU)};
\node[box, below=0.8cm of s2, xshift=1.55cm, text width=2.3cm] (S) {Summary of previous chunk};
\node[box, above=0.6cm of s2, text width=3.0cm, minimum height=0.8cm] (llm) {LLM (one JSON response)};
\draw[arr] (doc) -- (s1);
\draw[arr] (s1) -- node[lab, above] {$c_1..c_n$} (s2);
\draw[arr] (s2) -- node[lab, above] {enriched} (s3);
\draw[arr] (s3) -- (out);
\draw[arr] ([xshift=-0.85cm]K.north) -- node[lab, left] {$K_{i-1}$} ([xshift=-1.0cm]s2.south);
\draw[arr] ([xshift=-0.5cm]s2.south) -- node[lab, right, pos=0.55] {$k_i$} ([xshift=0.15cm]K.north);
\draw[arr] ([xshift=0.5cm]s2.south) -- node[lab, left, pos=0.55] {$s_i$} ([xshift=-0.35cm]S.north);
\draw[arr] ([xshift=0.65cm]S.north) -- node[lab, right] {$s_{i-1}$} ([xshift=1.0cm]s2.south);
\draw[arr] ([xshift=-0.5cm]s2.north) -- node[lab, left] {prompt} ([xshift=-0.5cm]llm.south);
\draw[arr] ([xshift=0.5cm]llm.south) -- node[lab, right] {fields} ([xshift=0.5cm]s2.north);
\end{tikzpicture}%
}
\caption{Indexing with \system{} on one document. Enrichment processes chunks in order. Call $i$ receives the summary $s_{i-1}$ of the previous chunk and, when the rolling key dictionary is enabled (arm E), the dictionary $K_{i-1}$; it returns key $k_i$ and summary $s_i$, and $K$ is reset per document. Dashed boxes are optional add-ons, tested in \S\ref{sec:mech}; related-key expansion acts at retrieval time and is not shown. Chunking and merging make no model calls.}
\label{fig:pipeline}
\end{figure*}

\subsection{Header-Led Chunking}
\label{sec:parsing}

\paragraph{Block parsing.}
After normalizing line endings, the parser reads the document line by line into blocks of eight types: YAML front matter (a \texttt{-{}-{}-} line at the start of the document, up to the next \texttt{-{}-{}-}); fenced code; header; table; list; blockquote; thematic break; and paragraph. Fences follow CommonMark (at least three backticks or tildes, indented at most three spaces, closed by a line of the same character at least as long; an unclosed fence runs to the end of the document). Headers are ATX (\texttt{\#} to \texttt{\#\#\#\#\#\#} followed by a space) or setext (a non-blank line underlined by \texttt{=} for level~1 or \texttt{-} for level~2). A table starts at a line containing \texttt{|} that is immediately followed by a delimiter row, and continues while lines are non-blank and contain \texttt{|}. A list starts at a bullet (\texttt{-}, \texttt{*}, \texttt{+}) or numbered item and continues over further items, indented lines, and blank lines that are followed by either. A blockquote is a run of lines starting with \texttt{>}. A paragraph ends at a blank line or at a line that starts a header, fence, list item, or blockquote, or that contains \texttt{|}. The parser does not handle raw HTML blocks or indented code.

A header stack tracks the section hierarchy: a header of level $\ell$ removes all entries of level $\geq \ell$ and is pushed. Every block records the \emph{section path}, the stack's titles joined by ``\texttt{ > }''. A chunk's section path (the prefix of arm S+T) is that of its first body block, or of its first header if it has no body.

\paragraph{Grouping.}
Blocks are grouped in document order with two thresholds, $\taumin = 100$ characters and $\taumax = 1{,}500$ characters (a soft limit). A chunk's size is the sum of its blocks' lengths. Let \emph{body} mean any non-header block. For each block:
\begin{enumerate}[label=(\roman*)]
  \item A header closes the current chunk if that chunk already has body. Consecutive headers with no body between them therefore stay together and open the next chunk.
  \item A body block closes the current chunk if that chunk already has body and adding the block would exceed $\taumax$. A chunk that holds only headers is never closed for size, so a header always travels with the start of its content.
  \item The block is appended. If it is a code block, table, YAML front matter, or blockquote and the chunk has reached $\taumax$, the chunk is closed.
\end{enumerate}
If the document ends with headers that have no body, they are appended to the last chunk. No block is ever split, so a single block longer than $\taumax$ becomes a chunk on its own.

After grouping, every chunk's body comes from a single section. A final pass then merges small chunks. Scanning in order, if the previous output chunk is shorter than $\taumin$ and the combined length is at most $2\taumax$, the next chunk is appended to it; this can repeat while the accumulated chunk stays below $\taumin$. A final chunk shorter than $\taumin$ is appended to its predecessor under the same cap. Joined texts are separated by a blank line, and the joined chunk takes the section path of its longer part. This pass is the only way a chunk can span sections.

\subsection{Enrichment and the Rolling Key Dictionary}
\label{sec:enrich}

Enrichment visits the chunks of a document in order and makes one LLM call per chunk. The prompt (Appendix~\ref{app:prompt}) contains the chunk's section path (or ``(no section)''), its position ``$i$ of $n$'', the previous chunk's summary (``(first chunk)'' for $i=1$; ``(unavailable)'' if the previous call produced no summary), the chunk text, and the rolling key dictionary rendered as one line per key, ``\texttt{- }\emph{key}\texttt{ (seen }$c$\texttt{x)}'', or as ``(none yet --- this is the first chunk)'' when it is empty. In arm E$-$K the dictionary is replaced by ``(not provided)''.

\paragraph{Requested fields.}
The model is asked for one JSON object with the seven fields of Table~\ref{tab:fields}. Title, summary, keywords and questions are indexed before the chunk text; entities, the key and the related keys are not indexed. The key drives key-based merging and the rolling dictionary, and the related keys drive related-key expansion.

\begin{table*}[!htbp]
\centering
\small
\begin{tabularx}{\linewidth}{@{}l l X X@{}}
\toprule
\textbf{Field} & \textbf{Type} & \textbf{Requested} & \textbf{Applied in code} \\
\midrule
\texttt{title}        & string   & 3--8 words & stripped; $\leq$1{,}000 chars \\
\texttt{summary}      & string   & 1--2 sentences, 30--60 words, what makes the chunk unique & stripped; $\leq$1{,}000 chars \\
\texttt{keywords}     & string[] & 5--8 specific, domain-relevant terms & a string is split on commas; non-strings dropped \\
\texttt{entities}     & object[] & name and type $\in$ \{PERSON, ORG, LOC, TECH, CONCEPT, EVENT, METRIC\} & items without a name dropped; missing type, or a bare string item, becomes CONCEPT \\
\texttt{questions}    & string[] & 2--3 natural questions the chunk answers & a string becomes a one-item list \\
\texttt{key}          & string   & specific subtopic, 2--5 words, lowercase; seven rules (see text) & stripped; $\leq$80 chars; lowercased \\
\texttt{related\_keys}& string[] & 0--3 rolling keys the chunk directly depends on & lowercased; any key not in $K_{i-1}$ dropped \\
\bottomrule
\end{tabularx}
\caption{Fields requested in the single enrichment call, as specified in the prompt, and the checks the code applies to the response.}
\label{tab:fields}
\end{table*}

\paragraph{Keys.}
The prompt defines the key as the specific subtopic that distinguishes the chunk within the document and gives seven rules (Listing~\ref{lst:prompt}): distinguish the chunk from others on the same broad topic; name the specific aspect covered; four example keys; share a key only if two chunks cover the same specific aspect and would read as one coherent piece when combined; reuse a key from the rolling dictionary if the chunk continues that discussion; do not use the document's broad topic; never use a single word that could describe the whole document. The code does not check these rules; it only lowercases, trims, and truncates the key. Keys are compared as exact strings after lowercasing, so near-synonyms are distinct keys. If a response cannot be parsed after the retries, the chunk keeps empty fields and an empty key.

\paragraph{Rolling key dictionary.}
$K$ maps each key to its first position, last position, and count. After call $i$ returns a non-empty key $k_i$, its entry is created with count 1, or its count is incremented and its last position set to $i$. If $K$ then holds more than $\Kmax = 40$ keys, only the 40 keys with the largest last position are kept (least-recently-used eviction), and $K$ is reordered by recency; otherwise keys are listed in insertion order. Only keys and counts are shown to the model. Because \texttt{related\_keys} is filtered against $K_{i-1}$, the dictionary before call $i$, a chunk cannot list its own new key as related. Algorithm~\ref{alg:enrich} gives the procedure.

\begin{algorithm*}[!htbp]
\caption{Enrichment with the rolling key dictionary (one document)}
\label{alg:enrich}
\begin{algorithmic}[1]
\Require Chunks $C=[c_1,\ldots,c_n]$ of one document; LLM $\mathcal{L}$; cap $\Kmax$
\State $K \gets$ empty ordered map \Comment{reset per document}
\For{$i \gets 1$ to $n$}
  \If{$i = 1$} $s \gets$ ``(first chunk)''
  \Else\ $s \gets c_{i-1}.\text{summary}$, or ``(unavailable)'' if empty
  \EndIf
  \State $R \gets \text{keys}(K)$ \Comment{$K_{i-1}$}
  \State $p \gets \textsc{FormatPrompt}(c_i.\text{section\_title},\, i,\, n,\, s,\, c_i.\text{text},\, \textsc{Render}(K))$
  \State $r \gets \textsc{CallJSON}(\mathcal{L}, p)$ \Comment{JSON output, parse retries; $\bot$ if unparsable}
  \If{$r$ is a JSON object}
    \State set title, summary, keywords, entities, questions from $r$ with type coercion (Table~\ref{tab:fields})
    \State $c_i.\text{key} \gets \text{lower}(\text{trim}(r.\text{key}))$
    \State $c_i.\text{related\_keys} \gets [\,\text{lower}(x) : x \in r.\text{related\_keys},\ \text{lower}(x) \in R\,]$
  \EndIf \Comment{otherwise all fields stay empty}
  \If{$c_i.\text{key} \neq$ ``''}
    \If{$c_i.\text{key} \in K$}
      \State $K[c_i.\text{key}].\text{count} \mathrel{+}= 1$;\quad $K[c_i.\text{key}].\text{last} \gets i$
    \Else
      \State $K[c_i.\text{key}] \gets (\text{first}{=}i,\ \text{last}{=}i,\ \text{count}{=}1)$
    \EndIf
    \If{$|K| > \Kmax$}
      \State $K \gets$ the $\Kmax$ entries with largest last, ordered by last descending
    \EndIf
  \EndIf
\EndFor
\State \Return $C$
\end{algorithmic}
\end{algorithm*}

\subsection{Key-Based Merging}
\label{sec:restructure}

Key-based merging (arm M; \texttt{merge\_by\_keys} in the package) makes no model call. It groups the chunks of one document by exact key, regardless of their distance in the document or their sections; groups are visited in order of their first chunk, and the chunks of a group in document order. Each group is packed greedily with next-fit: a chunk joins the current bin if the bin's size plus the chunk's length plus 2 (for the separator) is at most $\taumerge = 3{,}000$ characters, and otherwise opens a new bin; earlier bins are not revisited. A chunk longer than $\taumerge$ stays on its own unchanged. Algorithm~\ref{alg:restructure} gives the procedure.

\paragraph{Merged chunk.}
A bin of two or more chunks becomes one chunk whose text is the fragments' texts in document order, separated by blank lines. Its section path, title, summary and key are those of the \emph{first} fragment, so its title and summary describe only that fragment. Keywords, questions and \texttt{related\_keys} are unions with duplicates removed (their order is not fixed across runs).

\paragraph{Chunks without a key.}
A chunk whose key is empty is never merged. If it is shorter than $\tauorphan = 200$ characters, lines of the form \texttt{[Section: \ldots]}, \texttt{[Previous: \ldots]}, and \texttt{[Next: \ldots]}, holding its section path and the summaries of its neighbours, are prepended to its text, which is then indexed with it. Finally, all output chunks are sorted by start line, so a merged chunk sits at its first fragment's position.

\begin{algorithm*}[!htbp]
\caption{Key-based merging (one document)}
\label{alg:restructure}
\begin{algorithmic}[1]
\Require Enriched chunks $C=[c_1,\ldots,c_n]$ of one document; $\taumerge$; $\tauorphan$
\State $G \gets$ ordered map from key to ascending chunk indices, over chunks with non-empty key
\State $O \gets$ indices of chunks with empty key
\State $C' \gets [\,]$
\For{each $(k, I)$ in $G$}
  \State $b \gets [I_1]$;\quad $z \gets |c_{I_1}.\text{text}|$
  \For{each $j$ in $I_2, I_3, \ldots$}
    \If{$z + |c_j.\text{text}| + 2 \leq \taumerge$}
      \State append $j$ to $b$;\quad $z \gets z + |c_j.\text{text}| + 2$
    \Else
      \State append $\textsc{Merge}(b)$ to $C'$;\quad $b \gets [j]$;\quad $z \gets |c_j.\text{text}|$
    \EndIf
  \EndFor
  \State append $\textsc{Merge}(b)$ to $C'$ \Comment{a one-chunk bin is returned unchanged}
\EndFor
\For{each $i \in O$}
  \If{$|c_i.\text{text}| < \tauorphan$}
    \State prepend section path and summaries of $c_{i-1}$, $c_{i+1}$ to $c_i.\text{text}$
  \EndIf
  \State append $c_i$ to $C'$
\EndFor
\State sort $C'$ by start line (stable)
\State \Return $C'$
\end{algorithmic}
\end{algorithm*}

\subsection{Related Keys}
Each chunk's \texttt{related\_keys} are a subset of the keys assigned earlier in the same document. Arm E+KX (\S\ref{sec:mech}) uses them at retrieval time only: after each retrieved chunk it inserts the not-yet-included chunks of the same document that share the chunk's key or are linked to it through related keys, until the token budget is filled.

\section{Enrichment Prompt}
\label{app:prompt}

Listing~\ref{lst:prompt} is the enrichment prompt sent for every chunk, copied from the \texttt{ENRICH\_PROMPT} template in \texttt{enricher.py}. The Python source doubles the literal braces of the JSON example; they are shown here as the model receives them. The six names in braces (\texttt{\{section\_title\}}, \texttt{\{position\}}, \texttt{\{total\}}, \texttt{\{prev\_summary\}}, \texttt{\{chunk\_text\}}, \texttt{\{rolling\_keys\}}) are substituted as described in Appendix~\ref{sec:enrich}: \texttt{\{rolling\_keys\}} becomes one line ``\texttt{- }\emph{key}\texttt{ (seen }$c$\texttt{x)}'' per key, or ``(none yet --- this is the first chunk)''; \texttt{\{prev\_summary\}} becomes the previous summary, ``(first chunk)'', or ``(unavailable)''; and \texttt{\{section\_title\}} becomes the section path or ``(no section)''. The prompt is sent as a single user message.

\begin{lstlisting}[language={},numbers=none,basicstyle=\ttfamily\scriptsize,columns=fullflexible,keepspaces=true,escapeinside={(*@}{@*)},caption={Enrichment prompt, verbatim.},label={lst:prompt}]
You are a document analysis expert. Analyze this text chunk from a Markdown document and extract structured metadata for a RAG (Retrieval-Augmented Generation) system.

**Section Path:** {section_title}
**Chunk Position:** {position} of {total} chunks
**Previous Chunk Summary:** {prev_summary}

**Chunk Text:**
{chunk_text}

**Rolling Keys (specific subtopics seen in previous chunks):**
{rolling_keys}

Extract the following in a single JSON response:

{
  "title": "A short descriptive title for this chunk (3-8 words)",
  "summary": "A 1-2 sentence summary (30-60 words) capturing the key information. Do NOT just repeat the first sentence. Focus on what makes this chunk UNIQUE (*@\textemdash@*) what would a search engine snippet show?",
  "keywords": ["5-8 salient terms or phrases, domain-specific preferred"],
  "entities": [
    {"name": "entity name", "type": "PERSON|ORG|LOC|TECH|CONCEPT|EVENT|METRIC"}
  ],
  "questions": ["2-3 specific questions this chunk can answer"],
  "key": "The SPECIFIC subtopic that makes this chunk UNIQUE within the document. 2-5 words, lowercase. CRITICAL RULES: (1) Must DISTINGUISH this chunk from other chunks about the same broad topic. (2) Think: if someone asked what SPECIFIC ASPECT this chunk covers, what would you say? (3) Examples: admissions process, gradient descent optimization, oauth token flow, q3 revenue breakdown. (4) Two chunks should share a key ONLY if they cover the EXACT same specific aspect and would make a coherent single piece when combined. (5) REUSE a key from the rolling keys list if this chunk CONTINUES the same specific discussion. (6) A key should NOT be the document broad topic (*@\textemdash@*) it must be more specific than that. (7) NEVER use a 1-word key that could describe the whole document.",
  "related_keys": ["From the rolling keys above, pick 0-3 keys that this chunk DIRECTLY discusses or depends on. Err on the side of fewer. Ask: would a reader need to read the related-key chunk to understand THIS chunk? If not, do not include it. An empty list is perfectly fine."]
}

Rules:
- "related_keys" must be a SUBSET of the rolling keys provided (*@\textemdash@*) only include genuinely relevant ones
- "entities" should include technical terms, proper nouns, and domain concepts with types: PERSON (people), ORG (organizations), LOC (locations), TECH (technologies/tools), CONCEPT (abstract concepts), EVENT (events/dates), METRIC (measurements/KPIs)
- "keywords" should be specific and domain-relevant (not generic words like "system", "data", "process")
- "questions" should be natural questions a user would ask that this chunk answers
- Return ONLY valid JSON, no extra text
\end{lstlisting}

\section{Earlier Evaluation of \system{}}
\label{app:withdrawn}

\Earlierversions{} evaluated an earlier version of the code with 30 queries over 18 Markdown documents and reported Recall@$k$ and MRR for fixed-size chunking, structural chunking, the full pipeline, and BM25 over structural chunks. An audit of the evaluation code and its stored outputs found the following; we treat those results as invalid.
\begin{itemize}
  \item The full-pipeline configuration embedded only the chunk text, so none of the generated metadata reached the retriever; it differed from structural chunking only through key merging, and its metrics differed from those of structural chunking on a single query.
  \item Texts were truncated to 900 characters before embedding, so a large share of the text of structural and merged chunks, but none of the 512-character chunks, was never embedded. Queries were embedded without the embedder's recommended query prefix.
  \item A chunk counted as relevant if it contained the first 60 characters of any hand-written gold string, case-insensitively. Several gold strings were common words that matched a large fraction of chunks, and the reported ``Recall@$k$'' was a hit rate.
  \item The corpus was described as the project's own documentation. It consisted of seven Wikipedia articles, six third-party README or cookbook pages, a copy of an earlier draft of the preprint, and four documents written by the preprint's author, one of them a synthetic survey; most queries could be answered only from those four documents. The queries and the gold strings were also hand-written.
  \item The paper stated that FAISS and \texttt{rank\_bm25} were used; the code used a NumPy cosine ranking and its own BM25 implementation.
  \item The merge statistics were misreported: key merging formed 23 merge groups from 48 chunks, removing 25 chunks, not ``13 groups, 28 chunks''.
  \item The configurations were run on slightly different versions of one corpus document, and the evaluation code and corpus were not released.
  \item No confidence intervals or statistical tests were reported. With 30 queries, one query moves a rate by 3.3 points, and the Wilson 95\% interval for 26 of 30 is [0.70, 0.95], so the reported differences between configurations are not distinguishable from one another.
\end{itemize}

\end{document}